\pdfoutput=1

\documentclass[11pt]{article}

\usepackage[]{acl}
\usepackage{multirow}
\usepackage{times}
\usepackage{latexsym}
\usepackage{graphicx}
\usepackage{booktabs}
\usepackage{longtable}
\renewcommand{\thefootnote}
{\fnsymbol{footnote}}
\usepackage[T1]{fontenc}

\usepackage[utf8]{inputenc}
\usepackage{graphicx}

\definecolor{y}{rgb}{1, 1, 0}
\usepackage{tcolorbox}
\newtcolorbox{important_y}{
    colframe=y!80,%
    colback=y!80,%
    left=1pt, right=1pt,%
    top=0.5pt, bottom=0.5pt,%
    boxsep=0pt,%
    hbox,
    before=\vspace{0em},
    after=\vspace{0em}
}

\usepackage{microtype}

\title{Return of EM: Entity-driven Answer Set Expansion for QA Evaluation}

\author{Dongryeol Lee\textsuperscript{1} \hspace{1cm} Minwoo Lee\textsuperscript{2$\ddagger$} \hspace{1cm} Kyungmin Min\textsuperscript{3} \\
{\bf Joonsuk Park\textsuperscript{4,5,6$\dagger$}} \hspace{1cm} 
{\bf Kyomin Jung\textsuperscript{1$\dagger$}}\\
  \textsuperscript{1}Dept. of ECE, Seoul National University, \textsuperscript{2}LG AI Research, \textsuperscript{3}IPAI, Seoul National University, \\
  \textsuperscript{4}NAVER AI Lab, 
  \textsuperscript{5}NAVER Cloud, 
  \textsuperscript{6}University of Richmond\\
  \texttt{\{drl123, kyungmin97, kjung\}@snu.ac.kr}\\ \texttt{minwoo.lee@lgresearch.ai} 
  \hspace{3mm}
  \texttt{park@joonsuk.org}\\}

\begin{document}
\maketitle

\footnotetext{\textsuperscript{$\dagger$} Corresponding authors.}
\footnotetext{\textsuperscript{$\ddagger$} Work done while he was in Seoul National University.}

\renewcommand*{\thefootnote}
{\arabic{footnote}}
\setcounter{footnote}{0}

\begin{abstract}
Recently, directly using large language models (LLMs) has been shown to be the most reliable method to evaluate QA models. 
However, it suffers from limited interpretability, high cost, and environmental harm. 
To address these, we propose to use soft exact match (EM) with entity-driven answer set expansion.
Our approach expands the gold answer set to include diverse surface forms, based on the observation that the surface forms often follow particular patterns depending on the entity type.
The experimental results show that our method outperforms traditional evaluation methods by a large margin.
Moreover, the reliability of our evaluation method is comparable to that of LLM-based ones, while offering the benefits of high interpretability and reduced environmental harm.\footnote{Code and datasets are available at \url{https://github.com/DongryeolLee96/ENTQA}.}
\end{abstract}



\section{Introduction}
The advancement of large language models (LLMs) has led to their use as QA models, resulting in answers in sentence form with increased lexical diversity. 
This evolution has made traditional lexical matching metrics like exact match (EM) and F1 score less effective in capturing the performance of these models~\citep{kamalloo-etal-2023-evaluating}. 
In response, there has been a growing trend of employing LLMs themselves as evaluators~\citep{adlakha2023evaluating, liu2023gpteval}, leveraging their extensive parametric knowledge. 
While LLMs have been shown to measure the performance of QA models more reliably, they lack interpretability,
often yielding unconvincing verdicts~\citep{wang2023evaluating}. 
Additionally, they incur considerable costs, which could prevent contributions from less-resourced research groups. 
Lastly, their heavy electricity consumption is raising concerns about the environmental harm~\citep{gowda2023watt,khowaja2023chatgpt}.

\begin{figure}[t]
\centering{
\includegraphics[width=\columnwidth]{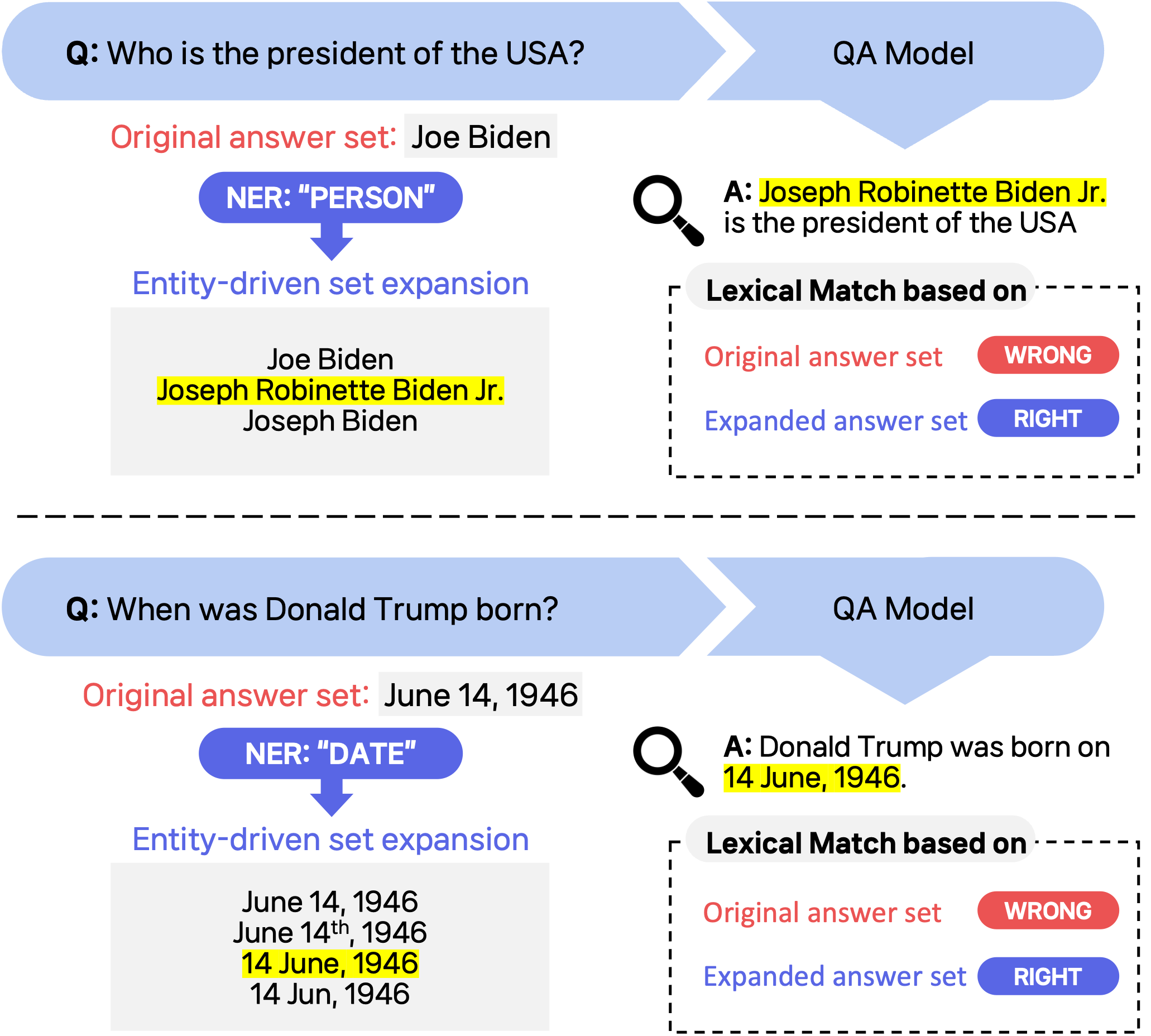}
}
\caption{Illustration of Our Method: We expand the original answer set based on the entity type, to include plausible surface forms for each entity type. 
By incorporating the Soft EM with \hspace{-0.05em}\raisebox{-0.8ex}{\begin{important_y}{expanded gold answer}\end{important_y}}, the QA model's prediction is correctly evaluated as right.
}
\label{fig:main}
\vspace*{-0.6cm}
\end{figure}

To address these limitations, we propose to use soft EM\footnote{Unlike EM, a candidate answer is marked correct if it contains a gold answer, even if they do not match exactly.} with entity-driven expansion of gold answers. 
Other means of expansion have been proposed, but they suffer from low reliability, measured as accuracy w.r.t. human verdict~\citep{si-etal-2021-whats}.
Our method is rooted in that surface forms of an answer can be diverse, but they often follow particular patterns depending on the entity type.
For example, \textit{Joe Biden} can also be referred to as \textit{Joseph Biden} or \textit{Joseph Robinette Biden Jr.}, while \textit{June 14, 1946} can be represented as \textit{14 June, 1946} or \textit{June 14th, 1946}, among others, as shown in Figure~\ref{fig:main}. 
We leverage the in-context learning abilities of LLMs, applying few-shot prompts specifically tailored for each entity type, to guide the expansion of the answer set. 
The use of soft EM significantly reduces the inference cost and environmental footprint, while making it explicit why a given candidate answer is correct or wrong.

We experimented with the outputs of five LLM-based QA models on two widely-used QA datasets---Natural Questions (NQ)~\citep{kwiatkowski2019natural} and TriviaQA (TQ)~\citep{joshi2017triviaqa}.
The results show that the \textbf{reliability} of our evaluation method is significantly higher than that of traditional lexical matching metrics---comparable to that of LLMs-based methods---while retaining the benefits of high \textbf{interpretability} and reduced \textbf{environmental footprint}.
More specifically, LLM-based methods require \textit{linearly increasing} inference calls (3,020 and 1,938) and costs (about \$0.50 and \$0.32) for evaluating each QA model in the NQ and TQ datasets, respectively. In contrast, our method requires a \textit{one-time} set of inference calls and a cost (\$1.93 for NQ; \$1.11 for TQ) for the initial expansion in each dataset.\footnote{Note, the expanded answer sets need to be shared, ideally along with the original dataset, for comparable evaluations across researchers. This in turn means the cost of expanding the answer set is a one-time cost \textit{for the whole community}, not individual researchers, drastically reducing the cost and environmental footprint.}

\section{Related Works}
Traditional QA models utilize a Retriever-Reader framework~\citep{karpukhin2020dense, chen2021salient, lewis2020retrieval, izacard2020leveraging, yu2021kg} to generate word-level answers, typically evaluated using lexical metrics such as Exact Match and F1. 
To address their limitation in recognizing semantic equivalence, answer set expansion methods utilizing knowledge bases like Freebase~\citep{bollacker2008freebase} and Wikipedia~\citep{si-etal-2021-whats} have been proposed. 
The rise of Large Language Models (LLMs) like InstructGPT~\citep{ouyang2022training} has shifted focus towards direct sentence-level answer generation, reducing reliance on external data sources~\citep{kamalloo-etal-2023-evaluating, roberts2020much, mallen2022not}. 
Additionally, newer model-based approaches for QA evaluation have emerged, employing fine-tuned models on answer equivalence datasets or using LLMs as evaluators to potentially enhance accuracy~\citep{bulian2022tomayto, risch2021semantic, vu2020ava, wang2023evaluating}.

\begin{table}[]
\centering
\resizebox{\columnwidth}{!}{%
\begin{tabular}{|l|c|l|}
\hline
Entity type &
  Format types &
  Examples \\ \hline
\multirow{4}{*}{\begin{tabular}[l]{@{}l@{}}\\ \\ \\\textbf{Numeric}\\- TIME\\- MONEY\\- QUANTITY\\- PERCENT\\- CARDINAL\\- DATE\\- ORDINAL\end{tabular}} &
  Numerals &
  \begin{tabular}[c]{@{}l@{}}\textbf{Q:} How many episodes are in season 2 of the handmades tale\\ \textbf{Gold Answer:} \textcolor{blue}{13}\\ \textbf{Model Prediction:} The Season 2 of the Handmaid's Tale\\ have \textcolor{red}{thirteen} episodes.\end{tabular} \\ \cline{2-3} 
 &
  \begin{tabular}[c]{@{}c@{}}Different\\ Representation\\ (symbols,\\ abbrev., order)\end{tabular} &
  \begin{tabular}[c]{@{}l@{}}\textbf{Q:} When was ye rishta kya kehlata hai started\\ \textbf{Gold Answer:} \textcolor{blue}{January 12, 2009}\\ \textbf{Model Prediction:} The Ye Rishta Kya Kehlata Hai started\\ in \textcolor{red}{12 Jan., 2009}.\end{tabular} \\ \cline{2-3} 
 &
  Specificity &
  \begin{tabular}[c]{@{}l@{}}\textbf{Q:} What's the population of fargo north dakota\\ \textbf{Gold Answer:} \textcolor{blue}{120,762}\\ \textbf{Model Prediction:} The population of Fargo, North Dakota is\\ \textcolor{red}{about 120,000}.\end{tabular} \\ \cline{2-3} 
 &
  Unit conversion &
  \begin{tabular}[c]{@{}l@{}}\textbf{Q:} How long is the movie son of god\\ \textbf{Gold Answer:} \textcolor{blue}{138 minutes}\\ \textbf{Model Prediction:} The movie Son of God is  \\\textcolor{red}{2 hours and 18 minutes} long.\end{tabular} \\ \hline
\multirow{2}{*}{\begin{tabular}[c]{@{}l@{}}\textbf{Non-numeric}\\ - PERSON\\ - GPE\\ - ORG\\ - Other\end{tabular}} &
  \begin{tabular}[c]{@{}c@{}}Different\\ representation\\ (symbols,\\ abbrev., order)\end{tabular} &
  \begin{tabular}[c]{@{}l@{}}\textbf{Q:} Where was the ncaa football championship game played 2018\\ \textbf{Gold Answer:} \textcolor{blue}{Atlanta, Georgia}\\ \textbf{Model Prediction:} The 2018 NCAA Football Championship\\ Game was played in \textcolor{red}{Atlanta, GA}.\end{tabular} \\ \cline{2-3} 
 &
  Specificity &
  \begin{tabular}[c]{@{}l@{}}\textbf{Q:} Who played lionel in all in the family\\ \textbf{Gold Answer:} \textcolor{blue}{Michael Evans}\\ \textbf{Model Prediction:} \textcolor{red}{Mike Evans} played Lionel Jefferson in All\\ in the Family.\end{tabular} \\ \hline
\textbf{N/A} &
  \begin{tabular}[c]{@{}l@{}}Contextual\\ Paraphrase\end{tabular} &
  \begin{tabular}[c]{@{}l@{}}\textbf{Q:} The pectoralis minor is located deep to which muscle\\ \textbf{Gold Answer:} \textcolor{blue}{beneath the pectoralis major}\\ \textbf{Model Prediction:} \textcolor{red}{under the pectoralis} \textcolor{red}{major muscle}\end{tabular} \\ \hline
\end{tabular}%
}
\vspace*{-0.2cm}
\caption{Categorization of surface form types depending on the entity types. Based on these surface forms, we sample a few-shot examples for each entity type.}
\vspace*{-0.5cm}
\label{tab:format_case_new}
\end{table}
\section{Entity-driven Answer Set Expansion}
\label{sec:method}
\subsection{Analysis of Surface Forms}
\label{sec:surface_form_analysis}
We begin by categorizing QA data based on the \textit{entity type} of their gold answers, employing Spacy's Named-Entity Recognizer (NER)\footnote{We utilized the "en\_core\_web\_lg" model from \url{https://spacy.io/}} to classify each entry into one of 19 categories – 18 predefined by Spacy and an additional N/A category for answers that do not conform to these classes. 
Based on answer categorizations, we generate answers using various QA models using training data from the NQ and TQ datasets.
Our co-authors then manually label each model's prediction according to its alignment with the original answer set. 
Through this process, we identify specific format patterns associated with different entity types. 
These patterns and their corresponding entity types are detailed in Table~\ref{tab:format_case_new}, providing insights into the variability of answer formats across entity categories.

Numeric entities, including TIME, MONEY, QUANTITY, PERCENT, CARDINAL, DATE, and ORDINAL, exhibit a diverse range of formats due to the varied expression of numeric values. 
For example, \textit{January 12, 2009} might be represented in different orders (e.g., \textit{12 January 2009}), abbreviations (e.g., \textit{Jan. 12, 2009}), or digit-to-text transformations forms (e.g., \textit{January 12th, 2009}), as illustrated in Table~\ref{tab:format_case_new}. 
These entities can also vary in units, such as \textit{138 minutes} expressed as \textit{2 hours and 18 minutes} or in abbreviated forms like \textit{138 mins} or \textit{2hrs and 18 mins}.
The N/A category, covering unique phrases or clauses, also exhibits significant paraphrasing variation.

Non-numeric entities like PERSON, GPE, and ORG, in contrast, demonstrate fewer variations. 
\textit{Mike Evans}, for instance, is an abbreviated form of \textit{Michael Evans} and can be expanded to a more specific form \textit{Michael Jonas Evans}, as depicted in Figure~\ref{tab:format_case_new}. 
Other entities, such as NORP, FAC, LOC, PRODUCT, EVENT, WORK\_OF\_ART, LAW, and LANGUAGE, usually exhibit minimal variation due to their lack of alternate forms.

\subsection{Answer Expansion Based on Entity Type}
\label{sec:expansion}
In line with our categorization, we implement an entity-specific strategy for answer set expansion. 
Our objective is to accurately enhance the gold answer set to reflect the typical format range associated with each entity type. 
We noted that distinct entity types exhibit varied answer formats, and certain expansions necessitate specific background knowledge. 
For example, understanding \textit{Mike Evans}'s full name or converting \textit{138 minutes} into hours and minutes is essential for proper expansion, as detailed in Table~\ref{tab:format_case_new}.

To address these challenges effectively, we utilize the parametric knowledge of InstructGPT\footnote{We also test the usage of Llama-2 in Appendix~\ref{app:llama2_exp}}, specifically its few-shot in-context learning feature. 
We choose eight illustrative examples per entity type from our training data, each accompanied by a manually expanded answer set that aligns with the format diversity of that entity type. 
The degree of expansion is carefully controlled by adjusting the number of expanded answers in these examples.
These selected examples are then utilized as few-shot prompts to facilitate the expansion of the original answer set. 


Details of our categorization and expansion approach are available in Appendix~\ref{app:entity}.

\section{Experiments}
\subsection{Dataset}
\label{sec:experiment_data}
We utilize two key datasets from the question-answering (QA) domain: Natural Questions (NQ)~\citep{kwiatkowski2019natural} and TriviaQA (TQ)~\citep{joshi2017triviaqa}. 
Specifically, we employed the EVOUNA dataset~\citep{wang2023evaluating}, which provides human judgment of answers, generated from these two datasets using five different QA models: DPR+FID~\citep{karpukhin2020dense, izacard2020leveraging}, GPT-3.5 (text-davinci-003), ChatGPT-3.5 (gpt3.5-turbo), ChatGPT-4, and BingChat~\citep{Bingchat2023}. 

\subsection{Evaluation Methods}
\label{sec:evaluation_method}
\paragraph{Model-based}
The \texttt{BEM} method, following \citet{bulian2022tomayto}, uses a pretrained BERT-base model~\citep{devlin2018bert} trained on answer equivalence datasets. 
Additionally, \texttt{Insteval} employs InstructGPT to evaluate prediction accuracy in relation to the question and reference answers, as per \citet{kamalloo-etal-2023-evaluating, wang2023evaluating}.

\paragraph{Lexical Matching-based}
The \texttt{Soft} and \texttt{Hard Exact Match (EM)}, mark a candidate answer as correct if it includes (\texttt{Soft}) or exactly matches (\texttt{Hard}) a reference answer. 
Additionally, an F1 score is used to measure the token overlap between the reference answer and prediction.

\paragraph{Soft EM with Answer Set Expansion}
Prior works have also explored answer expansion in different contexts.
They use either \texttt{Freebase}~\citep{si-etal-2021-whats}, for NQ and TQ datasets or \texttt{Wikipedia} for the TQ dataset~\citep{joshi2017triviaqa}\footnote{We utilized the Wikipedia-based expansion version of TriviaQA as released by the original authors. However, there is no publicly available code to adapt this method to the NQ.}. 
Our InstructGPT-based expansions employ variants like \texttt{Inst-zero} (only-instruction), \texttt{Inst-random} (randomly selected few-shot), and \texttt{Inst-entity} (entity type-specific few-shot as detailed in Section~\ref{sec:method}).
Based on the expanded answer set, soft EM is used to assess the model predictions.

The accuracy of these methods is determined by comparison against human annotations.


\subsection{Results \& Analysis}
\paragraph{Reliability}
Table~\ref{tab:against_hum} presents the accuracy of various evaluation metrics with respect to human judgment tested on the output of five different QA models. 
In particular, our soft EM with entity-driven answer set expansion metric (Inst-entity) is consistently more reliable than other lexical match-based ones with or without answer set expansion.
These metrics' effectiveness is notably reduced when applied to the output of LLM-based QA models, primarily due to LLMs' tendency to generate answers in sentences, which is not the case with ours.

\begin{table}[]
\centering
\resizebox{\columnwidth}{!}{%
\begin{tabular}{lcccccc}
\toprule[1.5pt]
\multicolumn{7}{c}{\textbf{Natural Questions}}                                                                                  \\ \hline
Evaluation Method       & FiD           & GPT3.5 & ChatGPT3.5 & ChatGPT4 & \multicolumn{1}{c|}{BingChat} & Avg.                 \\ \hline
\multicolumn{6}{l|}{{Model-based}}                                                                & \multicolumn{1}{l}{} \\
BEM            & \underline {93.5}    & 73.6   & 77.9       & 82.1     & \multicolumn{1}{c|}{84.0}     & 82.2                 \\
Insteval  & 91.8          & \underline {85.2}    & \textbf{86.2} & \textbf{89.2} & \multicolumn{1}{c|}{\textbf{88.0}} & \textbf{88.1} \\ \hline
\multicolumn{6}{l|}{{Lexical Matching-based}}                                                          & \multicolumn{1}{l}{} \\
Soft EM        & 89.7          & 84.9   & 80.5       & 82.9     & \multicolumn{1}{c|}{82.7}     & 84.1                 \\
Hard EM         & 86.9          & 37.3   & 28.5       & 21.2     & \multicolumn{1}{c|}{20.1}     & 38.8                 \\
F1             & \textbf{94.4} & 40.2   & 31.5       & 23.4     & \multicolumn{1}{c|}{20.5}     & 42.0                 \\ \hline
\multicolumn{6}{l|}{{Soft EM with Answer Set expansion}}                                           & \multicolumn{1}{l}{} \\
Freebase            & 89.8          & 85.5   & 81.7       & 83.9     & \multicolumn{1}{c|}{83.9}     & 85.0                 \\
Inst-zero      & 85.4          & 79.4   & 79.3       & 82.0     & \multicolumn{1}{c|}{83.8}     & 82.0                 \\
Inst-random & 88.1          & 83.8   & 82.2       & 86.0     & \multicolumn{1}{c|}{86.6}     & 85.3                 \\
\textbf{Inst-entity (Ours)}  & 91.0          & \textbf{86.8} & \underline {85.7}    & \underline {88.2}    & \multicolumn{1}{c|}{\underline {87.7}}    & \underline{87.9} \\ \hline \hline
\multicolumn{7}{c}{\textbf{TriviaQA}}                                                                                           \\ \hline
Evaluation Method       & FiD           & GPT3.5 & ChatGPT3.5 & ChatGPT4 & \multicolumn{1}{c|}{BingChat} & Avg.                 \\ \hline
\multicolumn{6}{l|}{{Model-based}}                                                                & \multicolumn{1}{l}{} \\
BEM            & \underline {93.8}    & 89.2   & 88.3       & 92.2     & \multicolumn{1}{c|}{90.3}     & 90.8                 \\
Insteval  & \textbf{96.4} & \textbf{94.2} & \textbf{94.9} & \textbf{96.0} & \multicolumn{1}{c|}{\textbf{95.1}} & \textbf{95.3} \\ \hline
\multicolumn{6}{l|}{{Lexical Matching-based}}                                                           & \multicolumn{1}{l}{} \\
Soft EM      & 88.0          & 87.5   & 87.3       & 86.2     & \multicolumn{1}{c|}{84.8}     & 86.8                 \\
Hard EM      & 85.3          & 40.8   & 22.0       & 13.2     & \multicolumn{1}{c|}{10.4}     & 34.3                 \\
F1             & 93.0          & 50.9   & 26.3       & 20.6     & \multicolumn{1}{c|}{10.6}     & 40.3                 \\ \hline
\multicolumn{6}{l|}{{Soft EM with Answer Set Expansion}}                                                        & \multicolumn{1}{l}{} \\
Freebase            & 90.6          & 89.4   & 89.0       & 88.4     & \multicolumn{1}{c|}{87.0}     & 88.9                 \\
Wiki           & 92.0          & 92.2   & 92.3       & 91.2     & \multicolumn{1}{c|}{90.1}     & 91.6                 \\
Inst-zero      & 88.1          & 86.1   & 88.6       & 89.7     & \multicolumn{1}{c|}{90.3}     & 88.6                 \\
Inst-random & 89.3          & 87.4   & 89.4       & 90.3     & \multicolumn{1}{c|}{91.2}     & 89.5                 \\
\textbf{Inst-entity (Ours)}  & 92.6          & \underline {92.5}    & \underline {93.3}    & \underline {93.0}    & \multicolumn{1}{c|}{\underline {92.4}}    & \underline{92.8} \\ 
\bottomrule[1.5pt]
\end{tabular}%
}
\vspace*{-0.1cm}
\caption{Reliability (accuracy w.r.t. human verdicts) of evaluation methods tested on the output of five QA models. \textbf{Bold} indicates the highest score, and \underline{underline} indicates the second highest score. For Lexical Matching-based and Model-based evaluations, the original gold answers from the respective datasets are used.}
\vspace*{-0.5cm}
\label{tab:against_hum}
\end{table}

\begin{figure}[t]
\centering{
\includegraphics[width=0.9 \columnwidth]{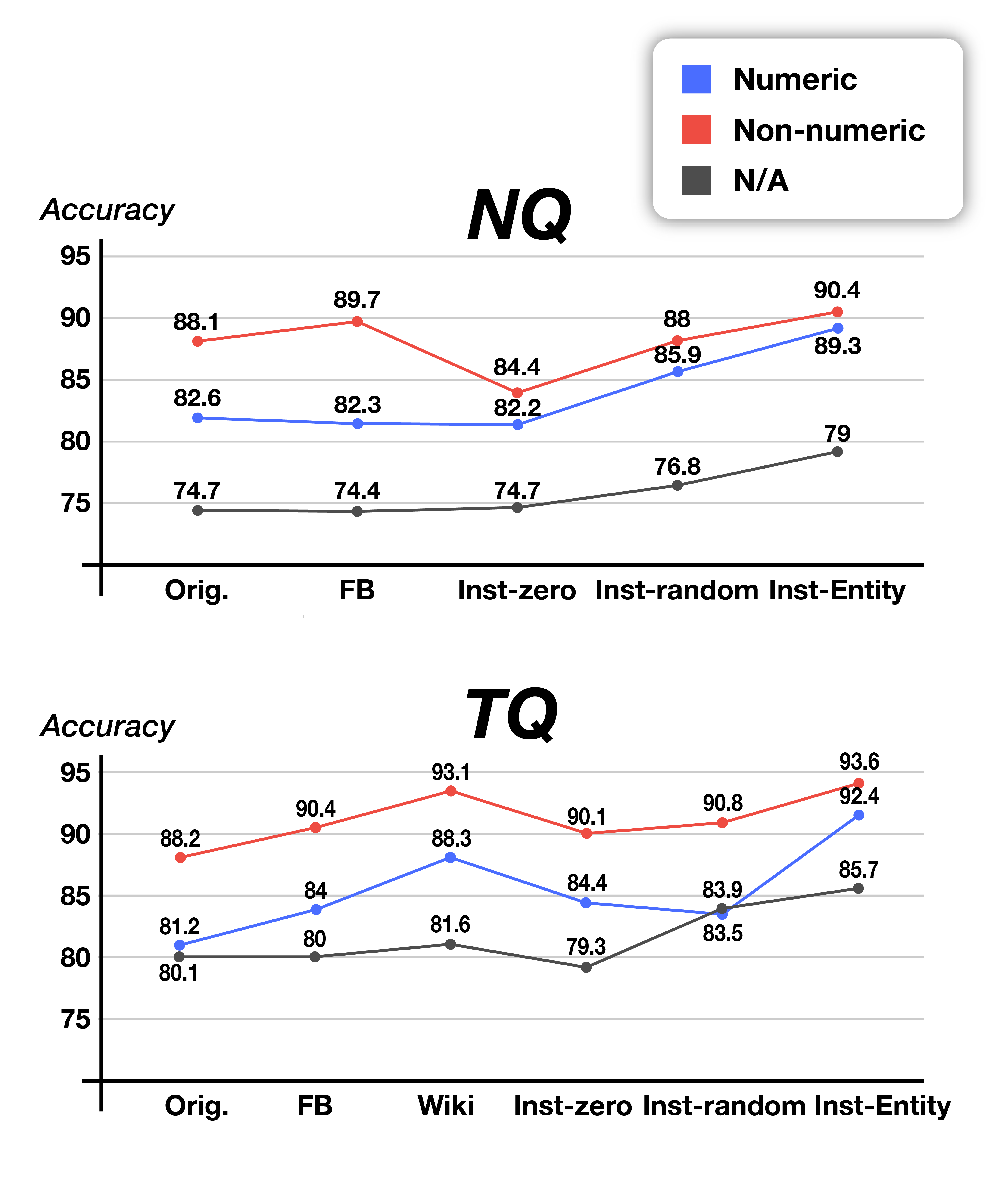}
}
\vspace*{-0.2cm}
\caption{Average accuracy against human labels across five QA models, using different answer set expansion methods. We separately report the accuracy based on entity types: Numeric, Non-numeric, and N/A.}
\vspace*{-0.3cm}
\label{fig:entity_result}
\end{figure}

Figure~\ref{fig:entity_result} details the specific accuracies of different expansion methods by entity type.
Numeric entities and the N/A category, which typically exhibit diverse surface forms, show lower accuracy with the original answer set. 
Our method significantly improves accuracy for these categories by effectively handling their diversity.
Among the answer set expansion baselines, Wikipedia-based expansion shows competitive performance in non-numeric entity types, leading to high performance in TQ since the majority of data (81.4\%) is non-numeric type. However, in numeric entity types, its performance is less effective. 
This can be attributed to the fact that Wikipedia entities are mostly related to non-numeric types, such as PERSON, LOC, and ORG.

\begin{figure}[t]
\centering{
\includegraphics[width=0.9\columnwidth]{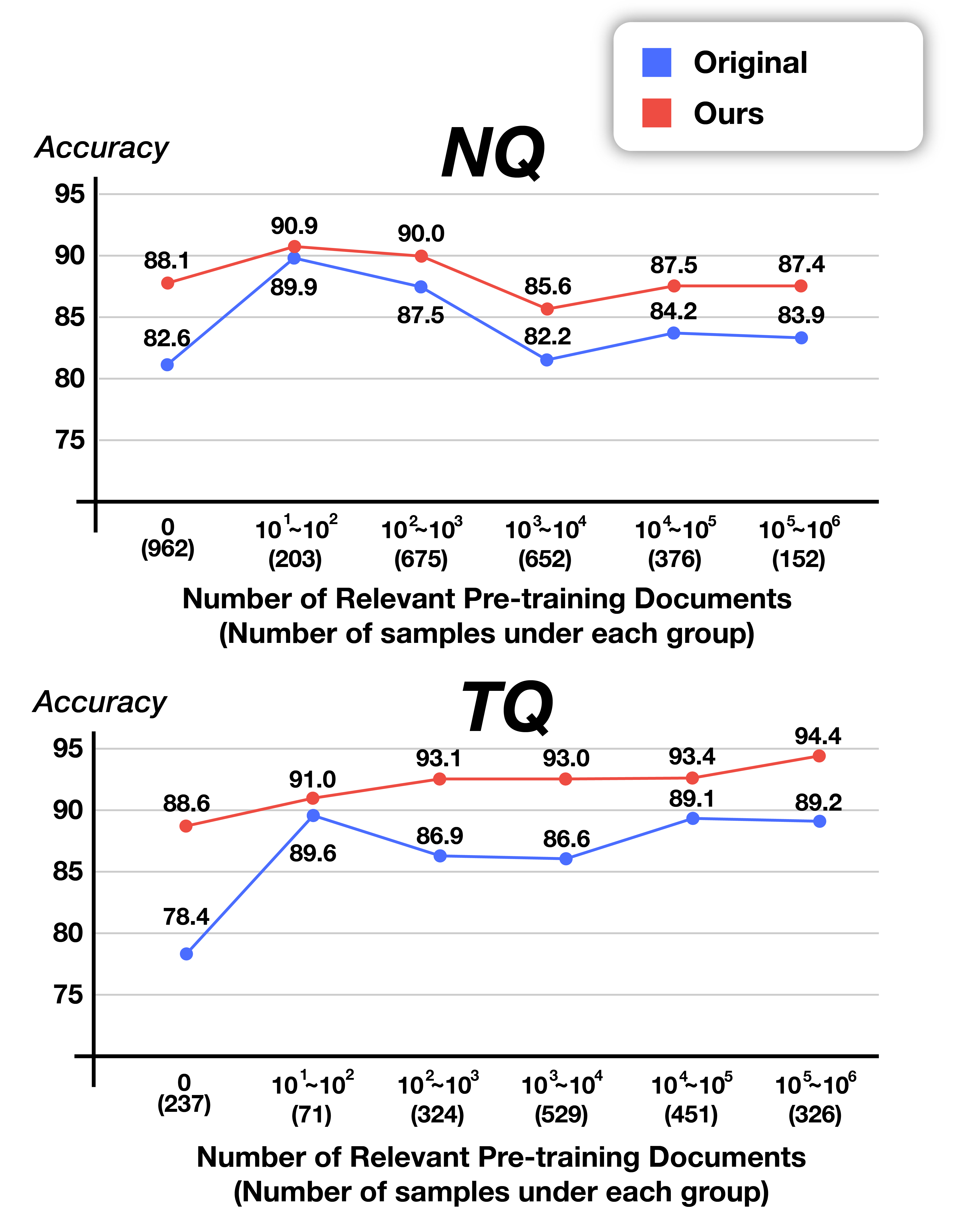}
}
\vspace*{-0.2cm}
\caption{Average accuracy against human labels across five QA models, using original answer set and our methods. We separately report the accuracy based on the rarity of the entity, which is measured by the number of its relevant docs in DBpedia~\citep{kandpal2023large}.}
\vspace*{-0.3cm}
\label{fig:rarity_result}
\end{figure}

The effectiveness of our method can vary depending on the entity's popularity~\citep{mallen2022not} or rarity~\citep{kandpal2023large}, as it relies on InstructGPT's background knowledge for each entity. 
Following ~\citet{kandpal2023large}, we also present the effectiveness of our method compared to Soft EM with the original answer set, segmented by the rarity of each entity. 
As shown in Figure~\ref{fig:rarity_result}, our method consistently maintains its effectiveness across varying levels of entity rarity.\footnote{Note that 0 relevant docs samples are usually numeric or N/A types answer entities which results in high accuracy gain.}


\paragraph{Interpretability}
While ours come close to the reliability of model-based metrics, the best model-based metric, Insteval, is still better correlated with human verdicts. 
However, a notable limitation of Insteval is the limited interpretability, operating as a black box and obscuring the logic behind its decisions~\citep{wang2023evaluating}. 
Table~\ref{tab:insteval_error} illustrates that 84\% of errors in Insteval are those without an understandable reason for the error.
For instance, it is difficult to fathom why Instaeval would mark “Jack Nicklaus [...]” as correct when the gold answer is “Gary Player”.
In contrast, our metric provides a rather clear justification: an answer is marked correct only if it contains a gold answer.

\begin{figure}[t]
\centering{
\includegraphics[width=1.0\columnwidth]{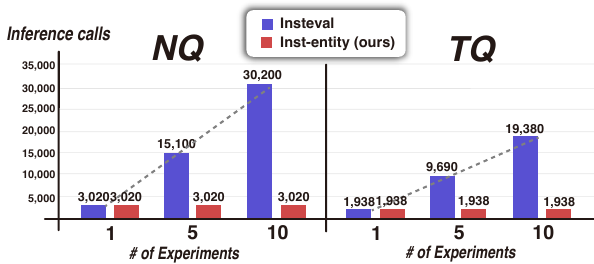}
}
\vspace*{-0.5cm}
\caption{Comparison of model-based evaluation and ours in terms of the inference calls. As the number of experiments increases, the inference calls for Insteval grow \textit{linearly}, whereas our method maintains a \textit{constant} number of interference calls.}
\vspace*{-0.3cm}
\label{fig:inference}
\end{figure}
\begin{table}[]
\centering
\resizebox{\columnwidth}{!}{%
\begin{tabular}{c|ll}
\toprule[1.5pt]
Type &
  Examples &
   \\ \cline{1-2}
\begin{tabular}[c]{@{}c@{}}Nonsensical \\Evaluation \\(84\%)\end{tabular}
 &
  \begin{tabular}[c]{@{}l@{}}\textbf{Question:}  who has played in the most masters tournaments \\ \textbf{Answer:}  {[}\textcolor{blue}{Gary Player}{]}\\  \textbf{Model prediction:}  \textcolor{red}{Jack Nicklaus} has played in the most \\Masters Tournaments, with a total of 44 appearances.\\ \\ \textbf{Human judgement} on \textbf{Model prediction:} Incorrect\\  \textbf{Insteval judgement} on \textbf{Model prediction:} Correct\end{tabular} &
   \\ \cline{1-2}
   \begin{tabular}[c]{@{}c@{}}Human \\Annotation \\Error \\(16\%) \end{tabular}
&
  \begin{tabular}[c]{@{}l@{}}\textbf{Question:}  who wins the final fight in real steel \\ \textbf{Answer:}  {[}\textcolor{blue}{Zeus}{]}\\  \textbf{Model prediction:}  The final fight in Real Steel is between \\Atom and Zeus. \textcolor{red}{Atom ultimately wins the fight}, becoming \\the reigning champion of the robot boxing world.\\ \\ \textbf{Human judgement} on \textbf{Model prediction:}  Correct\\ \textbf{Insteval judgement} on \textbf{Model prediction:} Incorrect\end{tabular} &
   \\ \bottomrule[1.5pt]
\end{tabular}%
}
\vspace*{-0.1cm}
\caption{Error instances from  Insteval, which do not match human judgement. The examples are taken from NQ and ten samples from each of the five QA models.}
\label{tab:insteval_error}
\vspace*{-0.5cm}
\end{table}

\paragraph{Environmental footprint}
Also, model-based metrics like Insteval require a significant number of inference calls, 3,020 for NQ and 1,938 for TQ, for each model being evaluated.
This means that to evaluate all five QA models in our experiments, 
a total of 15,100 inference calls were made for the NQ dataset, and 9,690 for the TQ dataset and would linearly grow w.r.t the number of experiments, as shown in Figure~\ref{fig:inference}.
In contrast, our metric does not require inference calls at the time of evaluation, but only when expanding the answer set initially---3,020 inference calls for NQ, and 1,938 for TQ. 
Since the expanded answer set can be made public, the community-wide implication is even larger: Our approach requires a fixed number of inference calls regardless of the number of researchers running experiments, but model-based metrics incur costs each time an evaluation takes place. 
Given the carbon emissions associated with the repetitive use of LLMs~\citep{patterson2021carbon, schwartz2020green}, this implies a substantial amount of carbon footprint. 
Also, our metric will help researchers with limited budgets, which is increasingly becoming a barrier in the age of LLMs~\citep{qin2023chatgpt, liu2023tcra, wang2023cost}.

Details of our experimental setup and additional experiments are available in Appendix~\ref{app:exp}.

\section{Conclusion}
We introduced a simple and effective approach for QA evaluation: Soft EM with Entity-driven answer set expansion. 
Our method demonstrated a competitive accuracy against human judgments, outperforming baseline evaluation metrics while maintaining simplicity, interpretability, and cost efficiency in terms of inference. 
This approach presents a viable solution for the QA evaluation challenges, particularly in the context of LLMs and the diverse range of answers they generate.

\section*{Limitations}
Our study relies on the Named-Entity Recognizer (NER) from Spacy, which may introduce errors in accurate named-entity tagging. 
Such inaccuracies could potentially affect the effectiveness of our entity-based answer set expansion. 
Developing a more precise NER tagger specifically for our experiments could mitigate this issue.

Furthermore, our analysis primarily focuses on errors arising from variations in surface forms of answers. 
However, there are instances where the data itself might be outdated, leading to expansions that produce not just varied forms of the gold answer but also entirely new, yet correct, answers. 
Addressing this aspect, including the development of methods to identify and handle outdated information, is left for future work.

\section*{Ethics Statement}
In our experiments, we employed the publicly available EVOUNA dataset~\citep{wang2023evaluating}, which is derived from the Natural Questions~\citep{kwiatkowski2019natural} and TriviaQA~\citep{joshi2017triviaqa} datasets. 
These datasets are widely recognized and used in the research community, ensuring the reliability and validity of our experimental data.

Furthermore, our use of the InstructGPT model for evaluating model predictions and expanding the answer set was conducted through OpenAI's official website\footnote{\url{https://openai.com/}}. 
All models employed in our experiments are sourced from publicly accessible platforms, including websites and GitHub repositories, adhering to open science principles.

\section*{Acknowledgements}
This work has been financially supported by SNU-NAVER Hyperscale AI Center.
This research was supported by the MSIT(Ministry of Science and ICT), Korea, under the ITRC(Information Technology Research Center) support program(IITP-2024-RS-2024-00437633) supervised by the IITP(Institute for Information \& Communications Technology Planning \& Evaluation).
This work was partly supported by Institute of Information \& communications Technology Planning \& Evaluation (IITP) grant funded by the Korea government(MSIT) [RS-2021-II211343, Artificial Intelligence Graduate School Program (Seoul National University) \& RS-2021-II212068, Artificial Intelligence Innovation Hub (Artificial Intelligence Institute, Seoul National University)], and the BK21 FOUR program of the Education and Research Program for Future ICT Pioneers, Seoul National University in 2024.
K. Jung is with ASRI, Seoul National University, Korea. The Institute of Engineering Research at Seoul National University provided research facilities for this work.

\bibliography{custom}

\appendix
\newpage
\section{Details of Entity-based expansion}
\label{app:entity}
In our approach to surface forms categorization in Section~\ref{sec:surface_form_analysis}, we utilize various QA models including LLAMA-2-13b~\citep{touvron2023llama}, and InstructGPT (gpt-3.5-turbo-instruct).
We also analyze the inference results from retrieve-then-read models (e.g. R2-D2) and end-to-end models (e.g. EMDR) which were provided in the previous work~\citep{kamalloo-etal-2023-evaluating}.

In our approach to answer set expansion in Section~\ref{sec:expansion}, we utilized a variation of InstructGPT~\citep{ouyang2022training}, specifically the gpt-3.5-turbo-instruct. 
We configured the hyperparameters of InstructGPT by setting the maximum output token length to 200 and the temperature parameter to 0. 
Additionally, the top\_p parameter was set to 1.

We conducted extensive experiments with a variety of instructions and selected the most effective instructions and few-shot examples in our pilot study.
The final prompt structure was created by concatenating these entity-type-based few-shot examples with instructions and the target answer set. 
The format of the prompt was as follows:

\textit{instruction, Question\_1, Expanded answer set\_1, ..., Question\_8, Expanded answer set\_8, Question\_target, Original answer set\_target}

``You are a given a question and a set of gold-standard reference answers (split with /) written by experts. Your task is to provide other forms of gold reference answers that can also be correct for the given question. Split your answers with /.'' was used for instruction.
Details of each example used in the prompts are described in Table \ref{tab:few_shot_nq1}, \ref{tab:few_shot_nq2}, \ref{tab:few_shot_tq1}, and \ref{tab:few_shot_tq2}.

\section{Experimental details and additional experiments}
\label{app:exp}
\subsection{Dataset details}
We utilized two major datasets for QA domains: Natural Questions~\citep{kwiatkowski2019natural} and TirviaQA~\citep{joshi2017triviaqa}.
Based on our entity type annotation, we report the statistics of these annotations for both datasets in Table~\ref{tab:data_stat}.
Additionally, we explore the impact of various answer set expansion methods on these datasets. 
Specifically, we report on how each method influenced the average number of answers per entity type.

\subsection{Surface accuracy}
One of the primary requirements for evaluation metrics is their ability to accurately capture the relative performance of different models. 
To assess this, we report surface accuracy as shown in Table~\ref{tab:surface_acc}, which represents the accuracy depicted by each metric when the model is evaluated.  
Furthermore, we evaluate surface accuracy through human judgment, which reflects how humans assess each model's performance.

Based on this human-derived surface accuracy, we ranked the QA models for the Natural Questions (NQ) and TriviaQA (TQ) datasets. 
The ranking outcomes were as follows: BingChat > ChatGPT4 > ChatGPT3.5 > FiD > GPT3.5 for NQ, and ChatGPT4 > BingChat > ChatGPT3.5 > FiD > GPT3.5 for TQ.

It is noteworthy that only our Inst-entity method and Freebase method aligned with the human-judged relative performance across both datasets. 
In contrast, the other metrics failed to accurately reflect the relative performance of the QA models.

\subsection{Additional experiments using Llama-2}
\label{app:llama2_exp}
In pursuit of a more cost-effective alternative to InstructGPT, we explored the performance of the widely utilized open-access model, Llama-2-chat-13b\footnote{\url{https://huggingface.co/meta-llama/Llama-2-13b-chat-hf}}.
Our investigation specifically examines the impact of substituting the LLMs used in both the model-based and answer set expansion methods detailed in Section~\ref{sec:evaluation_method}.
For the model-based method, we replaced Insteval's use of InstructGPT with Llama-2-chat-13b, resulting in the variants named Llama2-eval.
Similarly, for the answer set expansion method, we substituted InstructGPT of Inst-entity with Llama-2-chat-13b, leading to the variants named Llama2-entity.

Table~\ref{tab:llama_experiment} shows a performance comparison when using different LLMs (InstructGPT, Llama-2-13b). Interestingly, across both datasets, our methods (Llama2-entity and Inst-entity) demonstrate competitive performance against LLM-based methods (Llama2-eval and Insteval). In the NQ dataset, Llama2-entity even achieves SOTA accuracy, underscoring the effectiveness of our method. In the TQ dataset, Llama2-eval ranks second highest in most QA models, except in Newbing. This phenomenon, similarly detected in previous work by~\citep{wang2023evaluating}, is attributed to BingChat answers containing extraneous information and unique formatting. In contrast, our methods (Llama2-entity and Inst-entity) show stable performance, underscoring the robustness of our approach.

It is important to note that while Llama-2 models are publicly available and free to use, the repetitive inference cost and time are non-negligible. Therefore, the efficiency of our method remains valid across different LLMs.

\subsection{Case study: The impact of answer set expansion}

We conducted a case study analyzing instances where our entity-driven expansion (Inst-entity) outperforms the original answer set. 
To achieve this, we identified 50 cases from the NQ dataset where soft EM with the original answer set against human judgment was incorrect; simultaneously, soft EM with the expanded answer set against human judgment was correct.
Among the 50 cases, 43 demonstrated improvement through the expansion of the surface form of the original answers (e.g., \textit{Shirley Mae Jones} to \textit{Shirley Jones}). 
An intriguing finding from the case study was that the remaining 7 cases were rectified by the LLM using parametric knowledge to expand into semantically matching words. 
In Table~\ref{tab:ic_c}, while observing the expansion of \textit{Sheev Palpatin} to \textit{Emperor Palpatine}, it became evident that the LLM adeptly leveraged parametric knowledge.
This analysis highlights the effectiveness of our method in not only covering diverse lexical formats but also incorporating parametric knowledge, leveraging background information, and employing semantic equivalences for improved performance.

We also conducted a case study examining scenarios where our expansion method exhibited inferior performance than the original answer set. 
We selected 50 cases from the NQ dataset for detailed examination by employing a vice-versa sampling approach compared to the previous analysis.
Among these 50 cases, 47 revealed flaws in our expansion method, primarily stemming from DATE entities.
For instance, when the original answer was \textit{September 19, 2017}, our method expanded it to \textit{Sep, 2017} and \textit{2017}. However, if the answer generated by the QA model was \textit{September 5, 2017}, our attempt to reduce specificity in the date format led to errors. 
This highlighted the need for improvements when the QA model's answer triggers hallucination in specific DATE entities.
Moreover, there was a tendency to rely more on parametric knowledge than on explicit instructions, resulting in cases where answers were expanded to encompass information unrelated to the original answer.
The remaining 3 cases were attributed to human annotation errors resulting from human oversight. 
More detailed examples are provided in Table~\ref{tab:ic_c}.

\subsection{Case study: Rationale behind Soft EM}
Although our method, which utilizes Soft EM within an expanded answer set, shows reliable performance in two well-known QA datasets--Natural Questions and TriviaQA--it is important to consider the potential for differing results in specialized domains.
In well-known datasets, where LLMs like GPT-3.5 might have encountered similar data during training, it is less likely to generate responses with significant contextual errors (e.g. ``Joe Biden is not the president of US``), which cannot be deemed correct by matching ``Joe Biden''.
However, the potential for such errors could increase in specialized domains.

To investigate the effectiveness of Soft EM in these specialized domains, we conduct an additional experiment using the SciQ dataset~\citet{welbl2017crowdsourcing}, known for its specialized science-based content. 
We utilized the InstructGPT model (gpt-3.5-turbo-instruct) to answer 100 randomly sampled questions from the SciQ dataset. 
Each response was manually verified to determine whether the correct entity was identified and whether it was placed within an accurate context.

Out of 100 questions manually analyzed, the QA model generated an incorrect answer containing the correct entity on only 2 questions.
These instances were labeled as incorrect by humans despite containing the gold answer because they listed multiple answers, including the correct one.
This high level of correlation with human judgment underscores the reliability of Soft EM as an evaluation metric, capable of effectively measuring QA performance across specialized domains. 

\begin{table*}[]
\centering
\tiny
\begin{tabular}{|ll|}
\hline
\multicolumn{2}{|l|}{NQ} \\ \hline
\multicolumn{1}{|l|}{Entity Type} &
  Few-shot examples \\ \hline
\multicolumn{1}{|l|}{DATE} &
  \begin{tabular}[c]{@{}l@{}}Question: when was ye rishta kya kehlata hai started\\ Gold Answers: January 12, 2009/Jan 2009/2009/Jan 12, 2009/Jan 12th, 2009/January 2009/12th January 2009/12 January, 2009\\ Question: when is sharknado 6 going to be released\\ Gold Answers: August 19, 2018/2018/August 2018/Aug 2018/August 19th, 2018/19 August 2018/19 Aug 2018\\ Question: when was the last time tampa bay was hit by a hurricane?\\ Gold Answers: 1921/1920s/early 1920s/in early 1920s/Oct 1921/October 1921\\ Question: when did mutiny on the bounty take place?\\ Gold Answers: 28 April 1789/1789/April 1789/Apr 1789/April 28th, 1789/April 28, 1789/28th April, 1789/late 1700s\\ Question: game of thrones season 7 release date wiki\\ Gold Answers: July 16, 2017/July 16th, 2017/2017/July 2017/Jul 2017/Jul 16 2017\\ Question: On what date did India gain its independence?\\ Gold Answers: 15 August 1947/1947/Aug 1947/August 1947/August 15 1947/August 15th 1947/Aug 15, 1947\\ Question: When did De Braose die?\\ Gold Answers: 1211/early 1200s\\ Question: when did the tv show star trek start?\\ Gold Answers: September 8, 1966/September 8th, 1966/1966/Sep 8, 1966/Sep 8th, 1966/September, 1966/Sep, 1966\end{tabular} \\ \hline
\multicolumn{1}{|l|}{CARDINAL} &
  \begin{tabular}[c]{@{}l@{}}Question: How many physicians did Namibia have in 2002?\\ Gold Answers: 598/almost 600/approximately 600/five hundred ninety eight/approx. 600/almost 600\\ Question: what's the population of fargo north dakota\\ Gold Answers: 120,762/one hundred twenty thousand, seven hundred sixty-two/about 120,000/120762/about one hudred twenty thousand\\ Question: How many miles long is Metrorail?\\ Gold Answers: 24.4/24.4 miles/24.4 miles long/about 24 miles/approximately 24 miles/24.4 mi\\ Question: how many times chennai super kings win in ipl\\ Gold Answers: 91/ninety-one/91 times/ninety-one times/over 90 times\\ Question: How many of the Roman military were involved in the Battle of Allia River?\\ Gold Answers: 15,000 troops/fifteen thousands/fifteen thousands troops/15000/15,000/About 15,000\\ Question: What is the highest street number in the Bronx?\\ Gold Answers: 263/two hundreds sixty-three\\ Question: how many cards are in the game loteria\\ Gold Answers: 54/fifty-four/fifty-four cards/54 cards/54 in total\\ Question: How many died trying to defend the province in Kaliningrad?\\ Gold Answers: 300,000/three hundred thousands/about 300,000/approximately 300,000/300000\end{tabular} \\ \hline
\multicolumn{1}{|l|}{QUANTITY} &
  \begin{tabular}[c]{@{}l@{}}Question: How tall was Napoleon in centimeters?\\ Gold Answers: 168 cm/1.68m/1.68 m/1.68 meters/168 centimeters/5.5 inches/5ft 6 inches/5ft 6 in/5feet 6 in/5feet 6 inches\\ Question: How tall was John?\\ Gold Answers: 5 ft 5 in/5 feet 5 inches/165cm/1.65m/1.65 meters\\ Question: How large is Lafayette Park?\\ Gold Answers: 78-acre/seventy-eight acre/about 80 acres/approximately 80 acres\\ Question: What is the range of average elevation in the Sichuan Basin?\\ Gold Answers: 2,000 to 3,500 meters/2km to 3.5km\\ Question: how fast can sound travel in a second\\ Gold Answers: 331.2 metres/approximately 331.2 meters/approximately 331.2 m/1,086 feet/1,086 feet per second/approximately 330 meters per second\\ Question: During daytime how high can the temperatures reach?\\ Gold Answers: 80 °C (176 °F)/80 degrees Celcius/80°C/80 °C/176 °F/176 degrees Fahrenheit\\ Question: how far is beaumont texas from the ocean\\ Gold Answers: 30 miles/30 mi/thirty miles/30 miles away/about 30 miles\\ Question: How fast was the processor on the new Macintosh llfx?\\ Gold Answers: 40 MHz/40 Mega-Hz/40 MegaHertz/forty MHz/40 MHz fast\end{tabular} \\ \hline
\multicolumn{1}{|l|}{MONEY} &
  \begin{tabular}[c]{@{}l@{}}Question: How much in deposits did account holders withdraw from IndyMac in late June 2008?\\ Gold Answers: \$1.55 billion/1.55 billion dollars/approximately \$1.6 billion/around \$1.6 billion/approximately 1.6 billion dollars\\ Question: How much revenue did Apple announce for Q2 2007?\\ Gold Answers: \$5.2 billion/5.2 billion dollars/approximately \$5 billion/approximately \$5.2 billion\\ Question: how much did the new tappan zee bridge cost\\ Gold Answers: \$3.9 billion/3.9 billion dollars/approximately \$4 billion\\ Question: how much money does the iditarod winner get\\ Gold Answers: \$69,000/69,000 dollars/about \$70,000\\ Question: In 2014, how much research funding did Northwestern receive?\\ Gold Answers: \$550 million/550 million dollars/about \$550 million\\ Question: how much interest does the uk pay on its national debt\\ Gold Answers: PS43 billion/£43 billion/43 billion pounds/forty-three billion pounds\\ Question: What was reportedly the high value of of loot that the Ganj-i-Sawai had?\\ Gold Answers: £600,000/600,000 pounds/approximately 600,000 pounds/approximately £600,000/about £600,000\\ Question: What was the price tag for the private jet Schwarzenegger bought in 1997?\\ Gold Answers: \$38 million/38 million dollars/about \$38 million\end{tabular} \\ \hline
\multicolumn{1}{|l|}{PERCENT} &
  \begin{tabular}[c]{@{}l@{}}Question: Today, Mexico accounts for what percentage of Mennonites in Latin America?\\ Gold Answers: 42\%/42 percents/forty-two percent/about 42\%\\ Question: who owns 50 percent of the worlds wealth\\ Gold Answers: the top 1\%/top 1\%/1\%/one percent/the top one percent\\ Question: how much of the world's maple syrup does canada produce\\ Gold Answers: 80 percent/80\%/around 80\%/four-fifth\\ Question: what is the alcohol content of red stripe beer\\ Gold Answers: 4.7\%/about 4.7\%/about 5\%/approximately 5\%\\ Question: how much of canada's gdp is oil\\ Gold Answers: 2.9\%/almost 3\%/about 3\%\\ Question: What percentage of Australia's cotton crop was GM in 2009?\\ Gold Answers: 95\%/95 percent/ninety-five percent/around 95\%/almost 95\%\\ Question: what is the highest unemployment rate ever in the united states\\ Gold Answers: 25\%/one quarter/almost one quarter/almost 25\%/25 percent\\ Question: How many women at BYU do missionary work?\\ Gold Answers: 33 percent/33\%/one-third/about one-third/more than 30\%\end{tabular} \\ \hline
\multicolumn{1}{|l|}{TIME} &
  \begin{tabular}[c]{@{}l@{}}Question: how long is the movie son of god\\ Gold Answers: 138 minutes/2hrs and 18 mins/2hrs and 18 minutes/about 140 mins/138 mins\\ Question: how long is the all i have show\\ Gold Answers: two hours/2 hrs/2 hours/two hrs/120 minutes/120 mins\\ Question: how long is a wwe nxt live event\\ Gold Answers: 50-51 minutes/about 50 mins/between 50-51 minutes long/almost 51 mins long\\ Question: what is the running time of the last jedi\\ Gold Answers: 152 minutes/2 hrs 32 mins/2 hours 32 minutes/152 mins/about 2.5 hrs/about 2.5 hours\\ Question: when is the show this is us on tv\\ Gold Answers: 9pm/nine o'clock/at 9 o'clock/21:00\\ Question: when does a baby take their first breath\\ Gold Answers: about 10 seconds after delivery/10 seconds/10 secs/about 10 secs\\ Question: how long is an episode of once upon a time\\ Gold Answers: 43 minutes/43 mins/almost 45 minutes/forty-three mins/forty-three minutes\\ Question: How long before wake time is the lowest temperature reached?\\ Gold Answers: two hours/2 hours/2 hrs/two hrs/about 2 hours before\end{tabular} \\ \hline
\end{tabular}
\caption{Few-shot examples used in our entity-driven answer set expansion for numeric entity type in NQ.}
\label{tab:few_shot_nq1}
\end{table*}

\begin{table*}[]
\centering
\tiny
\begin{tabular}{|ll|}
\hline
\multicolumn{2}{|l|}{NQ} \\ \hline
\multicolumn{1}{|l|}{Entity Type} &
  Few-shot examples \\ \hline
\multicolumn{1}{|l|}{PERSON} &
  \begin{tabular}[c]{@{}l@{}}Question: who plays the bad guy in fifth element\\ Gold Answers: Gary Oldman/Gary L. Oldman/Gary Leonard Oldman/Gary/Oldman\\ Question: who does tess end up with on mcleods daughters\\ Gold Answers: Nick/Ryan/Nick Ryan\\ Question: who played mario in the super mario movie\\ Gold Answers: Bob Hoskins/Hoskins/Robert Hoskins/Robert William Hoskins/Robert W. Hoskins\\ Question: who holds the record for eating hot dogs\\ Gold Answers: Takeru Kobayashi/Kobayashi/Takeru "Tsunami" Kobayashi/Kobayashi Takeru\\ Question: who has played chad dimera on days of our lives\\ Gold Answers: Billy Flynn/Casey Jon Deidrick/William Flynn/Casey Deidrick/Casey J. Deidrick\\ Question: who ran the fastest 40 time in nfl history\\ Gold Answers: Bo Jackson/Vincent Edward "Bo" Jackson/Jackson\\ Question: who does vin diesel play in fast and furious 6\\ Gold Answers: Dominic Toretto/Torreto/Dominic "Dom" Toretto\\ Question: who played hey girl on have gun will travel\\ Gold Answers: Lisa Lu/Lisa Lu Yan/Lu\end{tabular} \\ \hline
\multicolumn{1}{|l|}{GPE} &
  \begin{tabular}[c]{@{}l@{}}Question: town replaced by kampala as ugandan capital in 1962\\ Gold Answers: Entebbe/Entebbe, Uganda\\ Question: which is the largest forest state in india\\ Gold Answers: Madhya Pradesh/Madhya Pradesh, India\\ Question: ranchi is capital of which state in india\\ Gold Answers: Jharkhand/Jharkhand, India\\ Question: where did kate and prince william get engaged\\ Gold Answers: Kenya/Rutundu, Kenya/Rutundu/East Africa\\ Question: where does tv show private eyes take place\\ Gold Answers: Toronto/Toronto, Canada/Toronto, Ontario/Ontario/Toronto, Ontario, Canada\\ Question: where is the netflix show the travelers filmed\\ Gold Answers: Vancouver, BC, Canada/Vancouver, BC/Vancouver, Canada/Canada/BC, Canada/Vancouver\\ Question: where is rhodochrosite found in the united states\\ Gold Answers: Colorado/Colorado, USA/Colorado, United States/Colorado state\\ Question: where was the ncaa football championship game played 2018\\ Gold Answers: Atlanta, Georgia/Georgia/Mercedes-Benz Stadium/Mercedes-Benz Stadium in Atlanta, Georgia/Atlanta\end{tabular} \\ \hline
\multicolumn{1}{|l|}{ORG} &
  \begin{tabular}[c]{@{}l@{}}Question: who has the most world series wins in mlb history\\ Gold Answers: New York Yankees/Yankees\\ Question: who did the vikings play in their first playoff game\\ Gold Answers: Atlanta/Atlanta Falcons/Falcons\\ Question: who was the publisher of brave new world\\ Gold Answers: Chatto \& Windus/Chatto and Windus/Chatto\&Windus\\ Question: who makes the fastest car in the world\\ Gold Answers: Bugatti/Bugatti automobiles/Bugatti automobiles S.A.S.\\ Question: where can you find naruto shippuden in english\\ Gold Answers: Neon Alley/on Neon Alley/in Neon Alley\\ Question: where is nanny mcphee and the big bang filmed\\ Gold Answers: University of London/Dunsfold Aerodrome/various London roads/Hambleden in Buckinghamshire/London/UK/Buckinghamshire\\ Question: who has the most shops in the uk\\ Gold Answers: Tesco/Tesco plc\\ Question: where does the majority of new york city's drinking water come from\\ Gold Answers: The Delaware Aqueduct/The Catskill Aqueduct/Catskill/Delaware\end{tabular} \\ \hline
\multicolumn{1}{|l|}{\begin{tabular}[c]{@{}l@{}}Other \\(NORP,\\ LOC,\\ WORK\_OF\_ART, \\FAC, \\PRODUCT, \\EVENT, \\LAW, \\LANGUAGE)\end{tabular}} &
  \begin{tabular}[c]{@{}l@{}}Question: where does the word coffee originally come from\\ Gold Answers: the Arabic qahwah/Arabic\\ Question: where can united states citizens find their civil liberties listed\\ Gold Answers: Bill of Rights/in Bill of Rights\\ Question: when was the salary cap introduced to the nhl\\ Gold Answers: During the Great Depression/Great Depression\\ Question: what kind of car does jay gatsby drive\\ Gold Answers: Rolls Royce/Rolls-Royce/Rolls-Royce 40\\ Question: elton john's first number one hit song\\ Gold Answers: "Crocodile Rock"/Crocodile Rock\\ Question: where does easy jet fly from in uk\\ Gold Answers: London Luton Airport/Luton Airport/London Luton\\ Question: what is the prison island in san francisco bay\\ Gold Answers: Alcatraz Island/Island Alcatraz\\ Question: what is the architectural style of the hagia sophia\\ Gold Answers: Byzantine/Byzantine empire|\end{tabular} \\ \hline
\multicolumn{1}{|l|}{Unknown} &
  \begin{tabular}[c]{@{}l@{}}Question: where did lucy jones come in the eurovision 2017\\ Gold Answers: 15th place/15th/fifteenth/fifteenth place\\ Question: How many physicians did Namibia have in 2002?\\ Gold Answers: 598/almost 600/approximately 600/five hundred ninety eight/approx. 600/almost 600\\ Question: how much of canada's gdp is oil\\ Gold Answers: 2.9\%/almost 3\%/about 3\%\\ Question: How tall was John?\\ Gold Answers: 5 ft 5 in/5 feet 5 inches/165cm/1.65m/1.65 meters\\ Question: how much money does the iditarod winner get\\ Gold Answers: \$69,000/69,000 dollars/about \$70,000\\ Question: who was the publisher of brave new world\\ Gold Answers: Chatto \& Windus/Chatto and Windus/Chatto\&Windus\\ Question: where did kate and prince william get engaged\\ Gold Answers: Kenya/Rutundu, Kenya/Rutundu/East Africa\\ Question: On what date did India gain its independence?\\ Gold Answers: 15 August 1947/1947/Aug 1947/August 1947/August 15 1947/August 15th 1947/Aug 15, 1947\end{tabular} \\ \hline
\end{tabular}
\caption{Few-shot examples used in our entity-driven answer set expansion for non-numeric entity type and N/A in NQ.}
\label{tab:few_shot_nq2}
\end{table*}
\begin{table*}[]
\centering
\tiny
\begin{tabular}{|ll|}
\hline
\multicolumn{2}{|c|}{TQ} \\ \hline
\multicolumn{1}{|l|}{Entity Type} &
  Few-shot examples \\ \hline
\multicolumn{1}{|l|}{DATE} &
  \begin{tabular}[c]{@{}l@{}}Question: The first Transit of Venus in the 21st century took place on 8 June 2004. \\What is the date of the next one?\\ Gold Answers: June 2012/2012 June 06/2012/June 6th, 2012/6 June 2012\\ Question: Forefathers Day is celebrated in the US on which date?\\ Gold Answers: 21 December/21th, December/December 21/Dec 21/December 21th\\ Question: In what year did Roald Amundsen reach the South Pole for the first time?\\ Gold Answers: 1911/14 December 1911/December 1911/December 14th, 1911/Dec 14th, 1911\\ Question: State of Israel is proclaimed.\\ Gold Answers: 1948/May 14, 1948/May, 1948/May 14th, 1948/14 May 1948\\ Question: An eruption in Iceland, known as the Laki eruption, where lava erupted from a 17-mile crack rather than from a standard volcano \\and lava tubes extended lava travel to more than 50 miles, devastated the country killing 80\% of livestock, caused starvation for over 20\% of \\the population, and affected areas as far as Africa and Asia. When was this?\\ Gold Answers: 1783-4/1783-1784/from 1783 to 1784\\ Question: In what year did 'Prohibition' officially end in America?\\ Gold Answers: 1933/December 5, 1933/Dec 5, 1933/December of 1933/December 5th, 1933\\ Question: Which date is Groundhog Day in the USA?\\ Gold Answers: February 2nd/Feb 2nd/February 2/Feb 2\\ Question: In which year was 'The Boston Tea Party'?\\ Gold Answers: 1773/December 16, 1773/December 1773/Dec 1773/Dec 16th, 1773/16 December 1773\end{tabular} \\ \hline
\multicolumn{1}{|l|}{CARDINAL} &
  \begin{tabular}[c]{@{}l@{}}Question: How many kilometres long is the walk - the longest race in men's athletics?\\ Gold Answers: 50/50km/fifty/fifty-kilometres\\ Question: "How many leagues did Captain Nemo travel ""under the sea""?"\\ Gold Answers: 20,000/20000/twenty thousand/twenty thousand leagues\\ Question: What is the maximum number of characters in a single SMS (text) message?\\ Gold Answers: 160/160 characters/one hundred sixty\\ Question: To the nearest 1000, what is the crowd capacity on Centre Court at Wimbiedon?\\ Gold Answers: 15,000/approximately 15,000/around 15,000/14,979/fifteen-thousands\\ Question: It's census time again. How many people did the US have in 1790 when the first census was taken?\\ Gold Answers: 4 million. 3,929,326, to be exact/3,929,326/around 4 million/4 million/almost 4,000,000\\ Question: On a standard dartboard, which number lies opposite 6?\\ Gold Answers: 11/eleven\\ Question: In the Washington Irving short story, for how many years did Rip van Winkle sleep in the Catskill Mountains?\\ Gold Answers: Twenty/20/Twenty years/20 year\\ Question: How long, to the nearest mile, is an Olympic marathon?\\ Gold Answers: 26/twenty six/approximately 26 miles/26 miles\end{tabular} \\ \hline
\multicolumn{1}{|l|}{QUANTITY} &
  \begin{tabular}[c]{@{}l@{}}Question: How tall is the monument 'Nelson's Column' in feet and inches?\\ Gold Answers: 170 feet and two inches/170 feet and 2 inches/170 ft 2 in/one-hundred seventy feet and two inches\\ Question: At which distance did Sebastian Coe win his Olympic gold medal in the Moscow games\\ Gold Answers: Fifteen hundred metres/1,500 m/1.5km/1.5 kilometres/one point five km\\ Question: How long is a volleyball court in feet?\\ Gold Answers: 60 feet/sixty feet\\ Question: In the Olympic shot put competition, what is the weight of the women's shot?\\ Gold Answers: 4 kilograms (8.82 lb)/4 kg/8.82 lb/4 kilograms/four kilograms/8.82 pounds\\ Question: What is the last event in the decathlon\\ Gold Answers: Fifteen hundred metres/1,500 metres/1.5km/0.93 miles/1.5 kilometres\\ Question: According to Dart Board Regulations, how high should the centre of the bullseye be from the floor in feet and inches?\\ Gold Answers: 5 feet 8 inches/5 ft 8 in/five feet eight inches\\ Question: To a thousand square miles, what is the area of New Jersey?\\ Gold Answers: 7,417 square miles/approximately 7,400 square miles/seven-thousands four-hundreds and seventeen square miles\\ Question: "In soccer, how far does ""the wall"" of players have to be from the spot where a free kick is to be taken?"\\ Gold Answers: 10 yards/9.144 meters/ten yards/9.144 m/30 feet/30 ft/360 inches\end{tabular} \\ \hline
\multicolumn{1}{|l|}{MONEY} &
  \begin{tabular}[c]{@{}l@{}}Question: If after spending 10\% of your money, you have \$180 left, how much did you start with?\\  Gold Answers: \$200/two-hundred dollars/200 dollars\\ Question: How much did Jerry Seinfeld reputedly turn down per episode when he refused to continue Seinfeld?\\ Gold Answers: \$5 million/5,000,000 dollars/five million dollars/\$5,000,000\\ Question: In dollars, how much did the 1997 film Titanic gross in its opening weekend in America?\\ Gold Answers: \$28,638,131/28,638,131 dollars/approximately \$29 million/almost \$29,000,000\\ Question: How much does it cost to buy Trafalgar Square on a monopoly board?\\ Gold Answers: £240/240 pounds/two-hundred forty pounds\\ Question: At 2013 what compensation had UK banks paid/set aside for the misselling of PPI (Payment Protection Insurance)?\\ Gold Answers: £18.4billion/18.4 billion pounds/£18,400,000,000/18,400,000,000 pounds\\ Question: It was announced in 2015 that Alexander Hamilton would be replaced on (What?), also called a sawbuck, alluding to the symbol X?\\ Gold Answers: \$10 bill/$10/$10 buck/ten bucks/ten dollars\\ Question: What does a colour TV licence cost?\\ Gold Answers: £145.50/145.50 pounds/approximately £145/almost £146\\ Question: In dollars, how much did the USA pay Russia for Alaskan territory in 1867?\\ Gold Answers: \$7,200,000/\$7.2 million/7.2 million dollars/7,200,000\end{tabular} \\ \hline
\multicolumn{1}{|l|}{PERCENT} &
  \begin{tabular}[c]{@{}l@{}}Question: An Ipsos MORI survey carried out this year showed politicians to have the lowest level of trust\\ of any occupation in the  U.K. What percentage of people trusted politicians\\ in general to tell the truth. ( accept within + or - 5 \% ) ?\\ Gold Answers: 18\%/eighteen percents/around 20\%/over 15\%\\ Question: (Up to) what degree of Neanderthal DNA is found in modern non-African people?\\ Gold Answers: 4\%/four percents/4 percents/four/up to 4\%\\ Question: In the United States, if liquor is defined as 80 proof, what is the percentage of alcohol by volume?\\ Gold Answers: 40\%/fourty percents/40 percents/40/two-fifth\\ Question: Seas and oceans make up roughly what proportion of the earth's surface?\\ Gold Answers: 70\%/seventy percents/approximately 70\%/around 70\%\\ Question: Twelve three-hundredths (12/300) expresssed as a percentage is?\\ Gold Answers: 4\%/four/4/four percent/one twenty-fifth\\ Question: What percentage of all Rolls-Royce Motor cars ever built are still roadworthy?\\ Gold Answers: Over 60\%/Over three-fifth/Over sixty percent/more than 60\%/above 60\%\\ Question: The human brain represents roughly what percentage of the body's resting metabolic rate (energy expended)?\\ Gold Answers: 20\%/one-fifth/twenty percent/approximately 20\%\\ Question: Approximately what percentage of Americans have appeared on television? 3\%, 11\% or 25\%?\\ Gold Answers: 25\%/one quarter/twenty-five percent/approximately 25\%\end{tabular} \\ \hline
\multicolumn{1}{|l|}{TIME} &
  \begin{tabular}[c]{@{}l@{}}Question: How long is the rest period between rounds in a professional boxing match?\\ Gold Answers: 60 seconds (one minute)/60 seconds/60 secs/one minute/one min./sixty seconds\\ Question: How long is a dog watch at sea?\\ Gold Answers: Two hours/2 hrs/2 hours/120 mins/120 minutes\\ Question: A snowflake takes approximately how long to fall fom sky to ground?\\ Gold Answers: One hour/1 hours/approximately 1 hours/60 minutes/60 min\\ Question: How long does a golfer get to find a lost ball\\ Gold Answers: Five minutes/5 minutes/5 mins/five mins\\ Question: How long is allowed between serves in an APT tennis match i.e. between 1st and 2nd serve?\\ Gold Answers: 20 SECONDS/twenty seconds/20 secs/20 seconds\\ Question: At what time of the day is the Ceremony of the Keys held in the Tower of London?\\ Gold Answers: 10pm/ten p.m./10 p.m./ten at night/10 at night\\ Question: Takuo Toda broke the world record for a paper plane flight, launched by hand from the ground, for what time?\\ Gold Answers: 26.1 seconds/around 26 seconds/approximately 26 secs/26.1 secs\\ Question: Because of the speed at which the earth and the moon move relative to the sun, a total solar eclipse can never last more than how long?\\ Gold Answers: 7 minutes 31 seconds/seven minutes thirty-one seconds/7 mins 31 secs/about 7.5 minutes\end{tabular} \\ \hline
\end{tabular}
\caption{Few-shot examples used in our entity-driven answer set expansion for numeric entity type in TQ.}
\label{tab:few_shot_tq1}
\end{table*}
\begin{table*}[]
\centering
\tiny
\begin{tabular}{|ll|}
\hline
\multicolumn{2}{|l|}{TQ}                            \\ \hline
\multicolumn{1}{|l|}{Entity Type} & Few-shot examples \\ \hline
\multicolumn{1}{|l|}{PERSON} &
  \begin{tabular}[c]{@{}l@{}}Question: Which French chef created Peach Melba in 1893?\\  Gold Answers: Auguste Escoffier/chef Auguste Escoffier/Georges Auguste Escoffier/Auguste/Escoffier\\ Question: Who managed England during the 1982 World Cup?\\ Gold Answers: RON GREENWOOD/Ronald Greenwood/Greenwood\\ Question: Donald Pleasance, Telly Savalas and Charles Gray have all played the role of which James Bond villain?\\ Gold Answers: Ernst Blofeld/Ernst S. Blofeld/Blofeld/Ernest\\ Question: What television host is married to Portia de Rossi?\\ Gold Answers: Ellen Degeneres/Ellen Lee Degeneres/Ellen L. Degeneres/Ellen\\ Question: Which World Heavyweight boxing champion-was known as 'The Cinderella Man'?\\ Gold Answers: JAMES BRADDOCK/JAMES J. BRADDOCK/James Walter Braddock\\ Question: In 1994 who became only the second actor to win successive Best Actor ‘Oscars’?\\ Gold Answers: Tom Hanks/Tom Jeffrey Hanks/Tom J. Hanks/Thomas Jeffrey Hanks/Thomas J. Hanks\\ Question: Who was William Shakespeare's mother\\ Gold Answers: Mary Arden/Mary Shakespeare/Mary\\ Question: What is the name of the top fashion designer who founder of the Fashion and Textile Museum in London?\\ Gold Answers: Zandra Rhodes/Dame Zandra Lindsey Rhodes/Zandra Lindsey Rhodes/Zandra L. Rhodes\end{tabular} \\ \hline
\multicolumn{1}{|l|}{GPE} &
  \begin{tabular}[c]{@{}l@{}}Question: What is the capital of Namibia?\\ Gold Answers: Windhoek/Windhoek, Namibia\\ Question: Where was the first commercial railway line built?\\ Gold Answers: Stockton to Darlington, UK/UK/Stockton, UK/Darlington, UK\\ Question: What is the Capital City of Latvia?\\ Gold Answers: Riga/Riga, Latvia\\ Question: Which country has the same name as a state of the USA?\\ Gold Answers: Western Georgia/Georgia\\ Question: In which Winter Olympics city did John Curry win gold in 1976?\\ Gold Answers: Innsbrück/InnsbruckInnsbruck, Austria\\ Question: By area, which is the largest state in the USA?\\ Gold Answers: Alaska/Alaska, United States/Alaska, USA\\ Question: Previously called Ezo/Yezo/Yeso/Yesso, what is Japan's north and second-largest island?\\ Gold Answers: Hokkaidou prefecture/Hokkaidou/Hokkaidou island\\ Question: The St Leger is run at which English racecourse?\\ Gold Answers: Doncaster, England/Doncaster\end{tabular} \\ \hline
\multicolumn{1}{|l|}{ORG} &
  \begin{tabular}[c]{@{}l@{}}Question: What organization won the 2012 Nobel Peace Prize?\\ Gold Answers: The European Union/EU\\ Question: What is the name of the bank in the UK television series ‘Dad’s Army’?\\ Gold Answers: Swallow Bank/Mainwaring's Bank\\ Question: Which car company made the Interceptor, ceasing production in 1976?\\ Gold Answers: JENSEN/JENSEN Motors\\ Question: Sam Walton founded which famous US retail chain in 1962?\\ Gold Answers: Walmart\\ Question: The original motto of which organisation was ‘Amidst War, Charity’?\\ Gold Answers: Red Cross/International Committee of the Red Cross/ICRC\\ Question: What magazine, with its iconic yellow border, was first published on Sept 22, 1888?\\ Gold Answers: National Geographic/National Geographic magazine\\ Question: Sony and Emirates Airlines withdrew their sponsorship in 2014 from which  global organization after ongoing corruption scandals?\\ Gold Answers: FIFA/Fédération Internationale de Football Association /FIFA (Fédération Internationale de Football Association)\\ Question: 'Core' is a brand of which computer technology company?\\ Gold Answers: Intel Corporation/Intel\end{tabular} \\ \hline
\multicolumn{1}{|l|}{\begin{tabular}[c]{@{}l@{}}Other \\(NORP,\\ LOC,\\ WORK\_OF\_ART, \\FAC, \\PRODUCT, \\EVENT, \\LAW, \\LANGUAGE)\end{tabular}} &
  \begin{tabular}[c]{@{}l@{}}Question: The vast majority of Indonesian people adhere to what religion?\\ Gold Answers: Islam/Islamic\\ Question: The island of Feurteventura lies in which body of water?\\ Gold Answers: Atlantic Ocean/Atlantic\\ Question: Which is the longest running Broadway musical in history?\\ Gold Answers: Phantom of the Opera/The Phantom of the Opera\\ Question: What is the world’s largest natural harbour?\\ Gold Answers: Sydney Harbour/Sydney Harbour\\ Question: In World War Two, which aircraft company manufactured the Stuka?\\ Gold Answers: Junkers/the junkers aircraft company\\ Question: What was first framed in 1864 and ratified in 1906 concerning the conduct of warfare?\\ Gold Answers: Geneva Convention\\ Question: What was the first US Federal statute to limit cartels and monopolies, passed in 1890, that still forms the basis for most antitrust litigation by \\the United States federal government?\\ Gold Answers: The Sherman Act\\ Question: Herbert Hoover and his wife Lou Henry Hoover often had public conversations in which language so that people could not eavesdrop on them?\\ Gold Answers: Mandarin Chinese/Mandarin\end{tabular} \\ \hline
\multicolumn{1}{|l|}{N/A} &
  \begin{tabular}[c]{@{}l@{}}Question: The2012 London Olympic Games were officially known as the games of what number Olympiad?\\ Gold Answers: 30th/thirtieth/30/thirty\\ Question: How many kilometres long is the walk - the longest race in men's athletics?\\ Gold Answers: 50/50km/fifty/fifty-kilometres\\ Question: Twelve three-hundredths (12/300) expresssed as a percentage is?\\ Gold Answers: 4\%/four/4/four percent/one twenty-fifth\\ Question: At which distance did Sebastian Coe win his Olympic gold medal in the Moscow games\\ Gold Answers: Fifteen hundred metres/1,500 m/1.5km/1.5 kilometres/one point five km\\ Question: How much does it cost to buy Trafalgar Square on a monopoly board?\\ Gold Answers: £240/240 pounds/two-hundred forty pounds\\ Question: Which car company made the Interceptor, ceasing production in 1976?\\ Gold Answers: JENSEN/JENSEN Motors\\ Question: Which country has the same name as a state of the USA?\\ Gold Answers: Western Georgia/Georgia\\ Question: In what year did 'Prohibition' officially end in America?\\ Gold Answers: 1933/December 5, 1933/Dec 5, 1933/December of 1933/December 5th, 1933\end{tabular} \\ \hline
\end{tabular}
\caption{Few-shot examples used in our entity-driven answer set expansion for non-numeric entity type and N/A in TQ.}
\label{tab:few_shot_tq2}
\end{table*}
\begin{table*}[]
\centering
\begin{tabular}{|l|cccc|ccccc|}
\hline
Dataset &
  \multicolumn{4}{c|}{Natural Questions} &
  \multicolumn{5}{c|}{TriviaQA} \\ \hline
 &
  \multicolumn{1}{c|}{} &
  \multicolumn{3}{c|}{avg. \# of gold ans.} &
  \multicolumn{1}{c|}{} &
  \multicolumn{4}{c|}{avg. \# of gold ans.} \\ \cline{3-5} \cline{7-10} 
\multirow{-2}{*}{Entity Type} &
  \multicolumn{1}{c|}{\multirow{-2}{*}{\#}} &
  \multicolumn{1}{c|}{Original} &
  \multicolumn{1}{c|}{Freebase} &
  Ours &
  \multicolumn{1}{c|}{\multirow{-2}{*}{\#}} &
  \multicolumn{1}{c|}{Original} &
  \multicolumn{1}{c|}{Freebase} &
  \multicolumn{1}{c|}{Wiki} &
  Ours \\ \hline
\textbf{DATE} &
  \multicolumn{1}{c|}{499} &
  \multicolumn{1}{c|}{1.7} &
  \multicolumn{1}{c|}{4.2} &
  15.2 &
  \multicolumn{1}{c|}{52} &
  \multicolumn{1}{c|}{1.0} &
  \multicolumn{1}{c|}{9.3} &
  \multicolumn{1}{c|}{6.7} &
  8.4 \\ \hline
\textbf{CARDINAL} &
  \multicolumn{1}{c|}{169} &
  \multicolumn{1}{c|}{1.6} &
  \multicolumn{1}{c|}{26.6} &
  6.7 &
  \multicolumn{1}{c|}{105} &
  \multicolumn{1}{c|}{1.0} &
  \multicolumn{1}{c|}{22.2} &
  \multicolumn{1}{c|}{5.9} &
  4.6 \\ \hline
\textbf{QUANTITY} &
  \multicolumn{1}{c|}{19} &
  \multicolumn{1}{c|}{1.8} &
  \multicolumn{1}{c|}{2.5} &
  15.8 &
  \multicolumn{1}{c|}{1} &
  \multicolumn{1}{c|}{1.0} &
  \multicolumn{1}{c|}{1.0} &
  \multicolumn{1}{c|}{1.0} &
  6.0 \\ \hline
\textbf{ORDINAL} &
  \multicolumn{1}{c|}{13} &
  \multicolumn{1}{c|}{2.2} &
  \multicolumn{1}{c|}{6.3} &
  9.0 &
  \multicolumn{1}{c|}{3} &
  \multicolumn{1}{c|}{1.0} &
  \multicolumn{1}{c|}{2.3} &
  \multicolumn{1}{c|}{9.3} &
  5.0 \\ \hline
\textbf{MONEY} &
  \multicolumn{1}{c|}{11} &
  \multicolumn{1}{c|}{1.3} &
  \multicolumn{1}{c|}{10.2} &
  4.5 &
  \multicolumn{1}{c|}{4} &
  \multicolumn{1}{c|}{1.0} &
  \multicolumn{1}{c|}{3.5} &
  \multicolumn{1}{c|}{1.8} &
  4.3 \\ \hline
\textbf{PERCENT} &
  \multicolumn{1}{c|}{10} &
  \multicolumn{1}{c|}{1.4} &
  \multicolumn{1}{c|}{24.5} &
  5.1 &
  \multicolumn{1}{c|}{2} &
  \multicolumn{1}{c|}{1.0} &
  \multicolumn{1}{c|}{23.0} &
  \multicolumn{1}{c|}{3.0} &
  5.0 \\ \hline
\textbf{TIME} &
  \multicolumn{1}{c|}{7} &
  \multicolumn{1}{c|}{1.1} &
  \multicolumn{1}{c|}{9.1} &
  6.3 &
  \multicolumn{1}{c|}{5} &
  \multicolumn{1}{c|}{1.0} &
  \multicolumn{1}{c|}{26.8} &
  \multicolumn{1}{c|}{12.6} &
  5.2 \\ \hline
\textbf{PERSON} &
  \multicolumn{1}{c|}{1035} &
  \multicolumn{1}{c|}{2.0} &
  \multicolumn{1}{c|}{13.4} &
  7.6 &
  \multicolumn{1}{c|}{744} &
  \multicolumn{1}{c|}{1.0} &
  \multicolumn{1}{c|}{12.9} &
  \multicolumn{1}{c|}{13.4} &
  5.6 \\ \hline
\textbf{GPE} &
  \multicolumn{1}{c|}{288} &
  \multicolumn{1}{c|}{2.2} &
  \multicolumn{1}{c|}{32.5} &
  9.6 &
  \multicolumn{1}{c|}{296} &
  \multicolumn{1}{c|}{1.0} &
  \multicolumn{1}{c|}{18.2} &
  \multicolumn{1}{c|}{27.3} &
  3.2 \\ \hline
\textbf{ORG} &
  \multicolumn{1}{c|}{198} &
  \multicolumn{1}{c|}{2.0} &
  \multicolumn{1}{c|}{11.8} &
  8.3 &
  \multicolumn{1}{c|}{380} &
  \multicolumn{1}{c|}{1.0} &
  \multicolumn{1}{c|}{18.3} &
  \multicolumn{1}{c|}{14.7} &
  2.4 \\ \hline
\textbf{NORP} &
  \multicolumn{1}{c|}{80} &
  \multicolumn{1}{c|}{1.8} &
  \multicolumn{1}{c|}{10.7} &
  4.2 &
  \multicolumn{1}{c|}{58} &
  \multicolumn{1}{c|}{1.0} &
  \multicolumn{1}{c|}{16.1} &
  \multicolumn{1}{c|}{12.8} &
  2.0 \\ \hline
\textbf{LOC} &
  \multicolumn{1}{c|}{46} &
  \multicolumn{1}{c|}{1.6} &
  \multicolumn{1}{c|}{6.1} &
  3.3 &
  \multicolumn{1}{c|}{32} &
  \multicolumn{1}{c|}{1.0} &
  \multicolumn{1}{c|}{9.4} &
  \multicolumn{1}{c|}{17.3} &
  2.0 \\ \hline
\textbf{WORK\_OF\_ART} &
  \multicolumn{1}{c|}{14} &
  \multicolumn{1}{c|}{1.9} &
  \multicolumn{1}{c|}{27.9} &
  5.7 &
  \multicolumn{1}{c|}{17} &
  \multicolumn{1}{c|}{1.0} &
  \multicolumn{1}{c|}{17.1} &
  \multicolumn{1}{c|}{6.9} &
  2.1 \\ \hline
\textbf{FAC} &
  \multicolumn{1}{c|}{17} &
  \multicolumn{1}{c|}{1.8} &
  \multicolumn{1}{c|}{6.6} &
  4.9 &
  \multicolumn{1}{c|}{17} &
  \multicolumn{1}{c|}{1.0} &
  \multicolumn{1}{c|}{5.4} &
  \multicolumn{1}{c|}{7.2} &
  2.0 \\ \hline
\textbf{PRODUCT} &
  \multicolumn{1}{c|}{13} &
  \multicolumn{1}{c|}{2.1} &
  \multicolumn{1}{c|}{5.9} &
  3.8 &
  \multicolumn{1}{c|}{21} &
  \multicolumn{1}{c|}{1.0} &
  \multicolumn{1}{c|}{16.8} &
  \multicolumn{1}{c|}{17.3} &
  2.1 \\ \hline
\textbf{EVENT} &
  \multicolumn{1}{c|}{8} &
  \multicolumn{1}{c|}{1.8} &
  \multicolumn{1}{c|}{5.8} &
  3.6 &
  \multicolumn{1}{c|}{9} &
  \multicolumn{1}{c|}{1.0} &
  \multicolumn{1}{c|}{5.3} &
  \multicolumn{1}{c|}{14.9} &
  2.0 \\ \hline
\textbf{LAW} &
  \multicolumn{1}{c|}{4} &
  \multicolumn{1}{c|}{1.8} &
  \multicolumn{1}{c|}{7.0} &
  3.8 &
  \multicolumn{1}{c|}{2} &
  \multicolumn{1}{c|}{1.0} &
  \multicolumn{1}{c|}{2.0} &
  \multicolumn{1}{c|}{5.0} &
  2.5 \\ \hline
\textbf{LANGUAGE} &
  \multicolumn{1}{c|}{3} &
  \multicolumn{1}{c|}{1.0} &
  \multicolumn{1}{c|}{10.0} &
  3.0 &
  \multicolumn{1}{c|}{1} &
  \multicolumn{1}{c|}{1.0} &
  \multicolumn{1}{c|}{1.0} &
  \multicolumn{1}{c|}{7.0} &
  2.0 \\ \hline
\textbf{Unknown} &
  \multicolumn{1}{c|}{586} &
  \multicolumn{1}{c|}{1.7} &
  \multicolumn{1}{c|}{14.9} &
  7.9 &
  \multicolumn{1}{c|}{189} &
  \multicolumn{1}{c|}{1.0} &
  \multicolumn{1}{c|}{10.2} &
  \multicolumn{1}{c|}{3.5} &
  5.2 \\ \hline
\textbf{Total} &
  \multicolumn{1}{c|}{3020} &
  \multicolumn{1}{c|}{1.8} &
  \multicolumn{1}{c|}{14.3} &
  8.9 &
  \multicolumn{1}{c|}{1938} &
  \multicolumn{1}{c|}{{ 1.0}} &
  \multicolumn{1}{c|}{{ 14.9}} &
  \multicolumn{1}{c|}{{14.1}} &
  {\color[HTML]{000000} 4.3} \\ \hline
\end{tabular}
\caption{Dataset Statistics for experiments. Avg. \# of gold and. denotes the average number of expanded gold answer sets for each answer set expansion method.}
\label{tab:data_stat}
\end{table*}

\begin{table*}[]
\renewcommand{\arraystretch}{1.0}
\centering
\small
\begin{tabular}{lcccccc}
\hline
\multicolumn{7}{c}{Natural Questions}                                                                            \\ \hline
Evaluation Method       & FiD             & GPT3.5          & ChatGPT3.5      & ChatGPT4        & BingChat        & Order \\ \hline
\multicolumn{7}{l}{Model-based}                                                                                  \\
BEM            & 70.1 (+1.2)  & 87.6 (+22.1) & 89.9 (+16.9) & 93.7 (+14.9) & 88.3 (+8.4)  & X     \\
Insteval  & 73.3 (+4.4)  & 76.2 (+10.7) & 82.3 (+9.3)  & 86.4 (+7.6)  & 87.2 (-7.3)  & X     \\ \hline
\multicolumn{7}{l}{Lexical Matching-based}                                                                             \\
Soft EM        & 59.1 (-9.8)  & 50.8 (-14.7) & 58.1 (-14.9) & 62.2 (-16.6) & 65.8 (-14.1) & X     \\
Hard EM        & 56.1 (-12.8) & 3.0 (-65.2)  & 2.0 (-72.8)  & 0.0 (-78.8)     & 0.0 (-79.9)     & X     \\
F1             & 64.1 (-4.8)  & 17.0 (-48.5) & 17.3 (-55.7) & 17.6 (-61.2) & 10.3 (-69.6) & X     \\ \hline
\multicolumn{7}{l}{Soft EM with Answer Set expansion}                                                                          \\
Freebase           & 60.0 (-8.9)  & 53.5 (-12.0) & 60.9 (-12.1) & 64.7 (-14.1) & 68.3 (-11.6) & O     \\
Inst-zero      & 69.1 (+0.2)  & 70.5 (+5.0)  & 75.9 (+2.9)  & 77.7 (-1.1)  & 79.0 (-0.9)  & X     \\
Inst-random & 68.5 (-0.4) & 68.2 (+2.7) & 75.1 (+2.1) & 77.5 (-1.3) & 79.2 (-0.7) & O \\
\textbf{Inst-entity (Ours)} & 67.4 (-1.5)  & 65.2 (-0.3)  & 72.6 (-0.4)  & 76.2 (-2.6)  & 77.7 (-2.2)  & O     \\ \hline
Human          & \textbf{68.9 (0)}    & \textbf{65.5 (0)}    & \textbf{73.0 (0)}     & \textbf{78.8 (0)}    & \textbf{79.9 (0)}    & O \\ \hline
\multicolumn{7}{c}{TriviaQA}                                                                                     \\ \hline
Evaluation Method       & FiD             & GPT3.5          & ChatGPT3.5      & ChatGPT4        & BingChat        & Order \\ \hline
\multicolumn{7}{l}{Model-based}                                                                                  \\
BEM            & 79.8 (-1.7)  & 85.0 (+6.6)  & 93.6 (+9.2)  & 95.3 (+5.1)  & 93.7 (+4.1)  & X     \\
Insteval  & 83.8 (+2.3)  & 82.4 (+4.0)  & 87.9 (+3.5)  & 92.8 (+2.6)  & 91.0 (+1.4)  & O     \\ \hline
\multicolumn{7}{l}{Lexical Matching-based}                                                                             \\
Soft EM        & 69.7 (-11.8) & 66.1 (-12.3) & 71.7 (-12.7) & 77.0 (-13.2) & 76.2 (-13.4) & O     \\
Hard EM        & 67.0 (-14.5) & 19.2 (-59.2) & 6.4 (-78.0) & 3.4 (-86.8) & 0.0 (-89.6)     & X     \\
F1             & 73.9 (-7.6)  & 36.0 (-42.4) & 25.1 (-59.3) & 25.9 (-64.3) & 7.3 (-82.3) & X     \\ \hline

\multicolumn{7}{l}{Soft EM with Answer Set expansion}                                                                          \\
Freebase       & 72.8 (-8.7)  & 69.2 (-9.2)  & 74.9 (-9.5)  & 79.9 (-10.3) & 79.5 (-10.1) & O     \\
Wiki           & 73.6 (-7.9)  & 70.9 (-7.5)  & 76.7 (-7.7)  & 82.2 (-8.0)  & 81.9 (-7.7)  & O     \\
Inst-zero      & 80.0 (-1.5)  & 82.6 (+4.2)  & 85.7 (+1.3)  & 89.2 (-1.0)  & 88.1 (-1.5)  & X     \\
Inst-random & 80.8 (-0.7)  & 82.9 (+4.5)  & 86.8 (+2.4)  & 90.5 (+0.3)  & 89.4 (-0.2)  & X     \\
\textbf{Inst-entity (Ours)} & 77.2 (-4.3)    & 75.9 (-2.5)  & 81.5 (-2.9)  & 86.5 (-3.7)  & 86.1 (-3.5)  & O     \\ \hline
Human          & \textbf{81.5 (0)}    & \textbf{78.4 (0)}    & \textbf{84.4 (0)}    & \textbf{90.2 (0)}    & \textbf{89.6 (0)}    & O \\ \hline
\end{tabular}%
\caption{Surface accuracy of each evaluation metric. The order indicates whether each evaluation metric reflects the relative performance order of the five QA models compared to human judgment.}
\label{tab:surface_acc}
\end{table*}
\begin{table*}[]
\centering
\begin{tabular}{lrrrrrr}
\hline
\multicolumn{7}{c}{Natural Questions} \\ \hline
Evaluation Method &
  \multicolumn{1}{l}{FID} &
  \multicolumn{1}{l}{GPT3.5} &
  \multicolumn{1}{l}{ChatGPT3.5} &
  \multicolumn{1}{l}{ChatGPT4} &
  \multicolumn{1}{l|}{BingChat} &
  \multicolumn{1}{l}{Avg.} \\ \hline
\multicolumn{6}{l|}{Model-based} &
  \multicolumn{1}{l}{} \\ 
Llama2-eval &
  88.2 &
  83.0 &
  83.4 &
  87.0 &
  \multicolumn{1}{r|}{82.5} &
  84.8 \\
Insteval &
  \textbf{91.8} &
  85.2 &
  \underline{86.2} &
  \textbf{89.2} &
  \multicolumn{1}{r|}{\underline{88.0}} &
  \underline{88.1} \\ \hline
\multicolumn{6}{l|}{Soft EM with Answer Set Expansion} &
  \multicolumn{1}{l}{} \\ 
Original &
  89.7 &
  84.9 &
  80.5 &
  82.9 &
  \multicolumn{1}{r|}{82.7} &
  84.1 \\
Llama2-entity &
  \underline{91.4} &
  \textbf{88.7} &
  \textbf{86.6} &
  \underline{89.0} &
  \multicolumn{1}{r|}{\textbf{88.7}} &
  \textbf{88.9} \\
Inst-entity &
  91.0 &
  \underline{86.8} &
  85.7 &
  88.2 &
  \multicolumn{1}{r|}{87.7} &
  87.9 \\ \hline
\multicolumn{7}{c}{TriviaQA} \\ \hline
Evaluation Method &
  \multicolumn{1}{l}{FID} &
  \multicolumn{1}{l}{GPT3.5} &
  \multicolumn{1}{l}{ChatGPT3.5} &
  \multicolumn{1}{l}{ChatGPT4} &
  \multicolumn{1}{l|}{BingChat} &
  \multicolumn{1}{l}{Avg.} \\ \hline\multicolumn{6}{l|}{Model-based} &
  \multicolumn{1}{l}{} \\
Llama2-eval  &
  \underline{94.6} &
  \underline{93.1} &
  \underline{94.3} &
  \underline{94.6} &
  \multicolumn{1}{r|}{{83.9}} &
  92.1 \\
Insteval&
  \textbf{96.4} &
  \textbf{94.2} &
  \textbf{94.9} &
  \textbf{96.0} &
  \multicolumn{1}{r|}{\textbf{95.1}} &
  \textbf{95.3} \\ \hline
\multicolumn{6}{l|}{Soft EM with Answer Set Expansion} &
  \multicolumn{1}{l}{} \\ 
Original &
  88.0 &
  87.5 &
  87.3 &
  86.2 &
  \multicolumn{1}{r|}{84.8} &
  86.8 \\
Llama2-entity  &
  91.5 &
  91.0 &
  91.4 &
  91.4 &
  \multicolumn{1}{r|}{90.0} &
  91.1 \\
  Inst-entity &
  92.6 &
  92.5 &
  93.3 &
  93.0 &
  \multicolumn{1}{r|}{\underline{92.4}} &
  \underline{92.8} \\\hline
\end{tabular}
\caption{Reliability (accuracy w.r.t. human verdicts) of evaluation methods using different LLMs tested on the output of five QA models. \textbf{Bold} indicates the highest score, and \underline{underline} indicates the second highest score. For Lexical Matching-based and Model-based evaluations, the original gold answers from the respective datasets are used. For Model-based evaluation, the original answer sets from Natural Questions and TriviaQA datasets are used.}
\label{tab:llama_experiment}
\end{table*}

\begin{table*}[]
\renewcommand{\arraystretch}{1.0}
\centering
\small
\begin{tabular}{ll}
\hline
\multicolumn{2}{c}{IC -\textgreater C}                     \\ \hline
\multicolumn{1}{c}{Type (\#)} & \multicolumn{1}{c}{Example} \\ \hline
Formatting (43) &
  \begin{tabular}[c]{@{}l@{}}Q: who played the mom in the partridge family\\ Original answer: {[}Shirley Mae Jones{]}\\ Expanded answer: {[}Shirley Mae Jones, Shirley Jones, \\Shirley J. Jones, Shirley Partridge, \\Shirley Renfrew Jones{]}\\ Model Prediction: Shirley Jones played the role of \\Shirley Partridge, the mom in the musical sitcom series \\"The Partridge Family"\end{tabular} \\ \hline
Background knowledge (7)      & \begin{tabular}[c]{@{}l@{}}Q: what was the emperor name in star wars\\ Original answer: {[}Darth Sidious, Sheev Palpatine{]}\\ Expanded answer: {[}Darth Sidious, Sheev Palpatine, Emperor Palpatine, \\Sheev, Emperor Sheev Palpatine{]}\\ Model Prediction: Emperor Palpatine\end{tabular}                            \\ \hline
\multicolumn{2}{c}{C-\textgreater IC}                       \\ \hline
\multicolumn{1}{c}{Type (\#)} & \multicolumn{1}{c}{Example} \\ \hline
Wrong Expansion (47) &
  \begin{tabular}[c]{@{}l@{}}Q: when is if loving you is wrong coming back season 4\\ Original answer: {[}September 19, 2017, March 7, 2018{]}\\ Expanded answer: {[}September 19, 2017, March 7, 2018, \\2017, 2018, Sep 19, 2017, Mar 7, 2018, \\Sep 2017, Mar 2018{]}\\ Model Prediction: Season 4 of the TV show "If Loving You \\Is Wrong" will premiere on OWN on \\Tuesday, September 5th, 2017.\end{tabular} \\ \hline
Human annotation error (3) &
  \begin{tabular}[c]{@{}l@{}}Q: a political leader during the roman empire was called\\ Original answer: {[}emperors{]}\\ Expanded answer: {[}emperors, Emperor, \\Roman Emperor, Roman leader, \\Roman political leader{]}\\ Model Prediction: Political leader during the Roman \\Empire: Such leaders were known by various titles \\depending on their role, including Emperor, Consul, and \\Senator, among others\end{tabular} \\ \hline
\end{tabular}
\caption{How can the expansion go Incorrect to Correct (IC->C) and can go Correct to Incorrect (C->IC). The examples are taken from NQ and ten samples from each five QA models.}
\label{tab:ic_c}
\end{table*}

\end{document}